\documentclass{article}

 \usepackage[eandd,final]{neurips_2026}

\usepackage[utf8]{inputenc} 
\usepackage[T1]{fontenc}    
\usepackage{hyperref}
\hypersetup{
    colorlinks=true,
    urlcolor=magenta 
}
\usepackage{url}            
\usepackage{booktabs}       
\usepackage{amsfonts}       
\usepackage{nicefrac}       
\usepackage{microtype}      
\usepackage{natbib}
\usepackage{subcaption}
\usepackage[table]{xcolor}
\usepackage{array}
\usepackage{graphicx}
\usepackage{adjustbox}
\usepackage{wrapfig}
\usepackage{framed}
\usepackage{makecell}
\usepackage{listings}
\usepackage{csquotes}
\usepackage{colortbl}
\usepackage{orcidlink}
\usepackage{multicol}
\usepackage{caption}
\usepackage{multirow}
\usepackage{fontawesome5}
\usepackage{longtable}
\usepackage{float}
\usepackage{arydshln}
\usepackage{pifont}         
\newcommand{\cmark}{\ding{51}}
\newcommand{\xmark}{\ding{55}}
\newcommand{\redxmark}{\textcolor{red}{\xmark}}
\newcommand{\greencmark}{\textcolor{green}{\cmark}}
\definecolor{deepgreen}{RGB}{0,100,0} 
\usepackage{xspace}
\newcommand{\eg}{\textit{e.g.}\@\xspace}
\newcommand{\ie}{\textit{i.e.}\@\xspace}
\definecolor{GainColor}{HTML}{3CB371}
\definecolor{LossColor}{HTML}{FF0000}  
\definecolor{ZeroColor}{HTML}{808080}  

\newcommand{\zero}[1]{$_{\color{ZeroColor}{+0.0}}$}  
\usepackage{pgf}

\title{VideoOdyssey: A Benchmark for Ultra-Long-Context and Omni-Modal Video Understanding}

\author{%
  Haichen He$^{1\ast}$, Jiayi Zhou$^{1\ast}$, Sifeng Shang$^1$, Yihan Hu$^3$, Yuanhan Zhang$^2$, Kaiyang Zhou$^1$ \\
  $^1$ Hong Kong Baptist University \\
  $^2$ S-Lab, Nanyang Technological University \\
  $^3$ GVC Lab, Great Bay University\\
  \url{https://videoodyssey-project.github.io/}
}

\begin{document}

\maketitle

\renewcommand{\thefootnote}{$\ast$}
\footnotetext{Equal contribution.}
\setcounter{footnote}{0}   
\renewcommand{\thefootnote}{\arabic{footnote}}

\begin{abstract}
\label{abstract}
Real-world long video understanding requires models to perform continuous tracking, information integration and memory retention over massive temporal spans within extreme video durations.  Mastering this intense cognitive load constitutes the fundamental bottleneck in long video understanding. While existing benchmarks have driven progress by scaling up video duration, their evaluation tasks often require comprehending only short and isolated video segments, falling short of capturing the challenge of ultra-long-context reasoning. To measure this cognitive load, we emphasize \textit{continuous certificate length}, defined as the video length a human must continuously watch to definitively answer a given question. Driven by this metric, we introduce \textbf{\textit{VideoOdyssey}}, a benchmark specifically designed for ultra-long-context and omni-modal video understanding. \textit{VideoOdyssey} is characterized by three key features: \textbf{1) Extreme video duration and diversity:} spanning 11 domains and 54 subcategories with an average video duration of 109 minutes; \textbf{2) Comprehensive evaluation scenarios:} offering two subsets to address different research focuses, i.e., \textit{VideoOdyssey-V} for probing the limits of visual understanding in MLLMs, and \textit{VideoOdyssey-AV} for evaluating synchronized audio-visual understanding for omni-modal models; \textbf{3) Ultra-long and multi-level continuous certificates:} extending the average continuous certificate to 16 minutes for \textit{VideoOdyssey-V} and 12.8 minutes for \textit{VideoOdyssey-AV}. Crucially, we design 5 granular levels from seconds to hours, providing a comprehensive diagnostic tool to evaluate models across varying context lengths and cognitive loads. Extensive evaluations show that bottlenecks of current MLLMs extend beyond simple retrieval to include struggles with continuous reasoning across varying context lengths, fine-grained perception, and non-verbal omni-modal understanding. We hope \textit{VideoOdyssey} will spur the development of next-generation MLLMs toward genuine real-world video understanding.

\end{abstract}   
\section{Introduction}
\label{intro}
Recent advancements in Multimodal Large Language Models (MLLMs) have pushed the boundaries of video understanding, facilitating a myriad of complex applications like autonomous driving and embodied AI. However, their real-world applicability remains untested. Authentic long video understanding requires models to perform continuous tracking, information integration, and memory retention over massive temporal spans within extreme video durations. This continuous high-density cognitive load is the true challenge of ultra-long-context reasoning in long video tasks.

While existing video benchmarks have made progress by scaling up raw video durations, their evaluation tasks often require understanding short and isolated video segments. This discrepancy stems from a severe annotation bottleneck: as video duration increases, the cognitive load required for humans to build logical chains and track continuous states grows exponentially. To circumvent this overwhelming difficulty, annotators instinctively compromise by labeling simple questions within narrow temporal windows. Consequently, simply extending the video duration falls short of reflecting the true difficulty of long video understanding.

\begin{figure}[t] 
\vspace{-0.3cm}
    \centering 
    \includegraphics[width=1.0\textwidth]{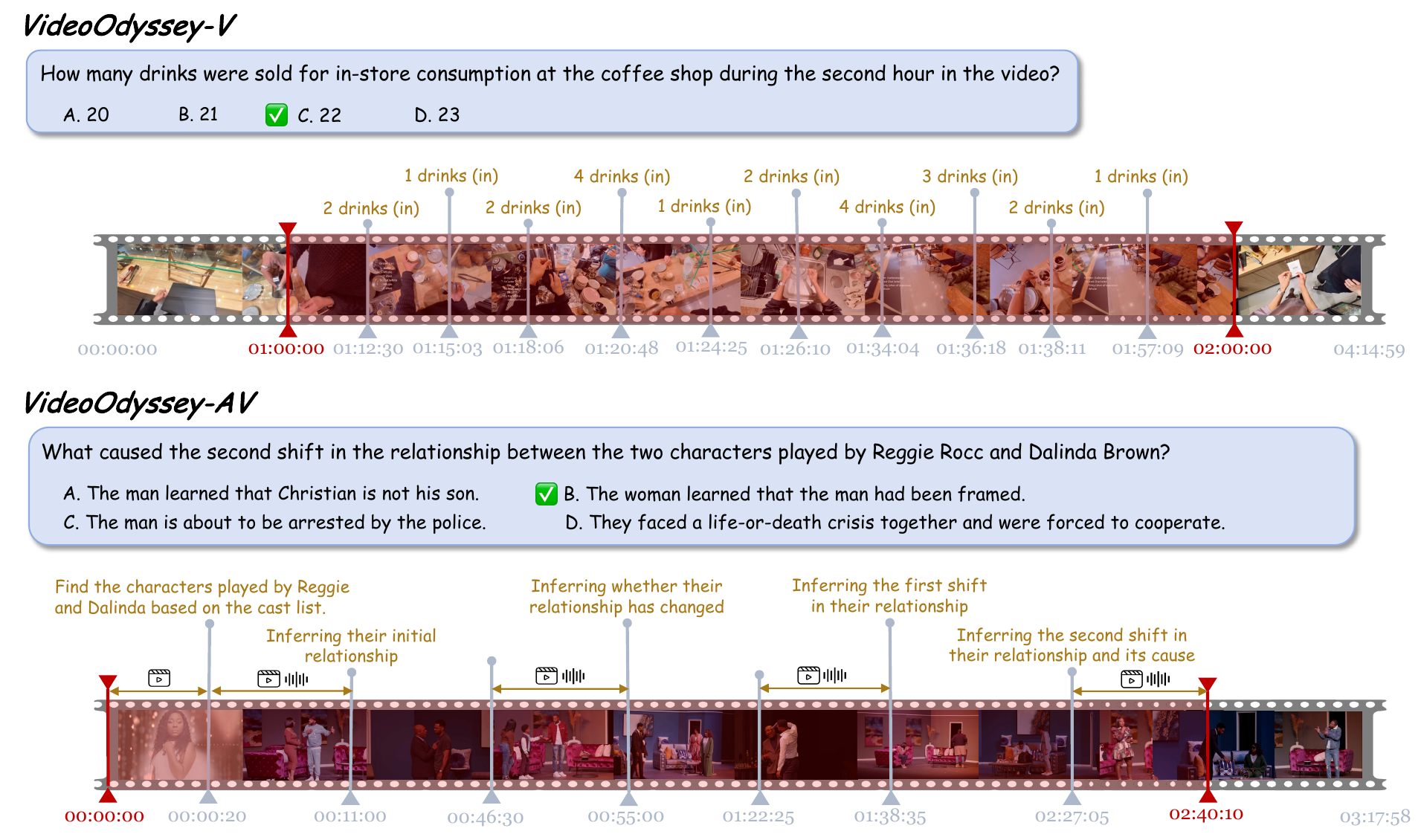} 
    \vspace{-0.5cm}
    \caption{Examples from our benchmark. In \textit{VideoOdyssey-V}, the model needs to consistently attend to detailed visual cues across an ultra-long time span, performing OCR-based counting tasks. In \textit{VideoOdyssey-AV}, the model needs to build a continuous logical chain of events over this massive time span, leveraging audio-visual cues to infer character relationships.}
    \label{fig:qa_example} 
    \vspace{-0.3cm}
\end{figure}

To explicitly measure this sustained cognitive load, we focus on \textit{continuous certificate length}, defined as the video length a human must continuously watch to answer a given question. This perspective builds upon the certificate length introduced in EgoSchema~\citep{mangalam2023egoschema}, which aggregates the total length of subclips needed to verify an answer. EgoSchema's formulation is highly effective for tasks where evidence is localized and isolated, such as action classification, detection, or simple video QA. However, as highlighted earlier, authentic long video understanding involves tasks that cannot be resolved merely by extracting isolated frames. For instance, consider a counting task in a surveillance scenario, such as \textit{``How many times did the man appear in this video?''}. To definitively answer this, humans cannot rely solely on the sparse moments the man is visible. 
They must invest unbroken attention across the entire long video to observe fine-grained details, track the target, and crucially, verify his absence at all other times. Such continuous tracking imposes a massive cognitive burden. The continuous certificate length explicitly quantifies this immense cognitive load induced by high-density, sustained attention.
We show examples of \textit{VideoOdyssey} demanding such extensive continuous certificate length in Fig. \ref{fig:qa_example}.

\begin{wrapfigure}{r}{0.48\textwidth} 
\vspace{-0.5cm}
  \centering
  \includegraphics[width=1.0\linewidth]{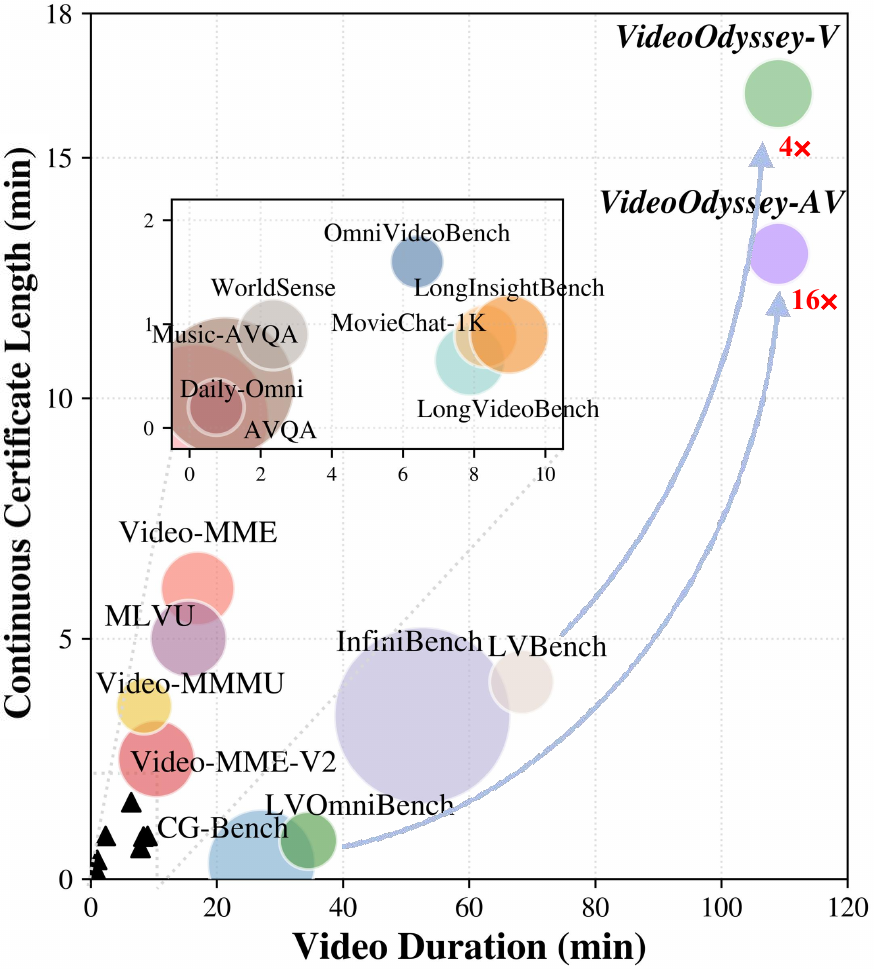} 
  \vspace{-0.5cm}
  \caption{Continuous certificate length across various video datasets.}
  \label{fig:dataset_bubble}
  \vspace{-0.2cm}
\end{wrapfigure}
Based on this metric, we introduce \textbf{\textit{VideoOdyssey}}, a pioneering benchmark specifically designed for ultra-long-context and omni-modal video understanding.  \textit{VideoOdyssey} features three key characteristics:
\textbf{1) Extreme video duration and domain diversity}: We collected 100 ultra-long videos from public platforms, spanning 11 domains and 54 fine-grained subcategories. The content ranges from structured narratives (\eg, \textit{TV, Movie}) to unstructured, complex content (\eg, \textit{Ego-centric videos, Surveillance}). The average video duration reaches 109 minutes. 
\textbf{2) Comprehensive evaluation scenarios and rigorous quality control}: We offer two specialized subsets to address different research focuses. \textit{VideoOdyssey-V} probes the limits of pure visual understanding in MLLMs across 14 tasks. Meanwhile, \textit{VideoOdyssey-AV} evaluates synchronized audio-visual understanding for omni-modal models across 18 tasks, incorporating three real-world audio types. Models are evaluated across multiple dimensions, including perception, cognition, summarization and temporal grounding.
Each subset is constructed through meticulous manual annotation and rigorous multi-stage quality control process.
\textbf{3) Ultra-long and multi-level continuous certificates}: We extend the average continuous certificate to an unprecedented 16 minutes for \textit{VideoOdyssey-V} and 12.8 minutes for \textit{VideoOdyssey-AV}. As shown in Fig.\ref{fig:dataset_bubble}, compared to the existing benchmarks with the longest video durations, we increase this metric by 4 times and 16 times for the pure visual and audio-visual domains, respectively. Crucially, we designed 5 granular continuous certificate levels ranging from seconds to hours. This design transforms \textit{VideoOdyssey} into the first comprehensive diagnostic tool capable of precisely tracing model performance across escalating cognitive loads, ultimately exposing critical bottlenecks to pave the way for genuine ultra-long-context reasoning.

Extensive benchmarking reveals staggering deficiencies in current MLLMs. Their capabilities are highly sensitive to the continuous certificate length, exposing fundamental comprehension flaws beyond mere retrieval difficulties. Specifically, models struggle with severe comprehension degradation in ultra-long contexts, while simultaneously failing to capture fine-grained details in ultra-short windows.
Furthermore, our analysis shows that RAG-based agentic approach also fails to bridge this gap. Its retrieval easily overlooks crucial visual details, while its discrete frame extraction inherently interrupts the event chains required for sustained tasks. 
Finally, current omni-modal integration remains heavily restricted to speech transcriptions, largely failing to comprehend non-verbal acoustic signals. Therefore, developing architectures capable of stable long-context reasoning, fine-grained perception, and non-verbal omni-modal understanding remains the critical next frontier.



To summarize, we have made the following contributions:

\begin{itemize}
    \item We introduce {\textit{VideoOdyssey}}, a pioneering benchmark designed for ultra-long-context and omni-modal understanding. Through rigorous manual annotation and strict quality control, we established two high-quality subsets: \textit{VideoOdyssey-V} for pure visual understanding and \textit{VideoOdyssey-AV} for synchronized audio-visual understanding.
    \item We use the continuous certificate length to explicitly quantify sustained cognitive load. Based on this metric, we design a multi-level diagnostic framework with 5 granular levels ranging from seconds to hours. This transforms \textit{VideoOdyssey} into the first comprehensive tool capable of precisely tracing model performance across escalating continuous contexts.
    \item We conduct extensive benchmarking of current state-of-the-art MLLMs and provide valuable insights. By highlighting fundamental limitations in ultra-long-context reasoning, fine-grained perception and omni-modal integration capabilities, our analysis offers actionable guidance for the development of next-generation intelligent systems.
\end{itemize}

\section{Related Work}
\label{related_work}
\paragraph{Multimodal Large Language Models}
The paradigm of video MLLMs has rapidly evolved from simple frame-wise feature aggregation toward more sophisticated temporal architectures. Early models primarily treated video as a sequence of individual images ~\citep{li2024llava, zhang2024llavanextvideo, liu2024llavanext, wang2025internvl3}. Recent models like Video-R1 ~\citep{feng2025video} and Video-KTR ~\cite{wang2026video} have recently emerged to resolve complex temporal logic through advanced optimization strategies  ~\citep{feng2025video, li2025videochat, tian2025ego, yan2025videochat, wang2026video}. Simultaneously, proprietary leaders such as Gemini-3.1-Pro ~\citep{google2025gemini3} has pushed the theoretical boundaries of sustained memory to million-token windows, enabling the processing of continuous, hour-long multimodal streams. Despite the progress, our benchmark indicates that current MLLMs are not good enough at dealing with ultra-long-context videos.

\paragraph{Long Video Benchmarks}
Early benchmarks primarily focused on short-form video clips ~\citep{Liu2024TempCompassDV, wu2021star_situated_reasoning, chen2023autoeval, li2023mvbench, xiao2021next, fang2024mmbench, mangalam2023egoschema}. As models evolved, evaluation shifted toward specialized domains and complex reasoning ~\citep{song2024moviechat, wu2024longvideobench, hu2025video, luo2025videoautoarena, zhou2025mlvu, fu2025video, chen2024cg, ataallah2025infinibench, wang2025lvbench}. 
Despite these advancements, as illustrated in the top section of Table \ref{tab:dataset_comparison}, a persistent trade-off between temporal scale and reasoning depth remains. 
Although LongVideoBench ~\citep{wu2024longvideobench} specifically targets long contexts, its average duration (7.88 minutes) and continuous certificate (0.7 minutes) remain limited. Even ``ultra-long'' efforts like InfiniBench ~\citep{ataallah2025infinibench} and LVBench ~\citep{wang2025lvbench}, which reach hour-long average durations, exhibit a relatively shallow reasoning depth, with average continuous certificate length of only 3.4 and 4.1 minutes, respectively. In contrast, \textit{VideoOdyssey-V} simultaneously maximizes both axes, pushing the boundaries to a 109-minute average duration and an unprecedented 16-minute average continuous certificate length. With multi-level certificate lengths, \textit{VideoOdyssey-V} provides a more rigorous and precise testbed for sustained long-context reasoning.

\paragraph{Audio-Visual Benchmarks}
Early audio-visual benchmarks were predominantly constrained to short-form clips ~\citep{yang2022avqa, li2022learning, geng2025longvale, zhou2025daily, li2025omnivideobench, han2025longinsightbench, hong2025worldsense} or static image-audio pairs ~\citep{li2024omnibench, gong2024av}. While these datasets spurred initial research into omni-modal integration, their limited temporal scales fall short of reflecting the complexity of the real world. Furthermore, as detailed in the bottom section of Table \ref{tab:dataset_comparison}, although recent effort LVOmniBench ~\citep{tao2026lvomnibench} scales the raw video durations to approximately 35 minutes, it suffers from a critical lack of reasoning depth, with a required context window of merely 0.8 minutes. Video-MME-v2 \citep{fu2026video} introduces audio-related tasks, but the conflation of modalities within its design makes it difficult to accurately assess models' true modal understanding capabilities. \textit{VideoOdyssey-AV} not only scales to a 109-minute average duration with a 12.8-minute average continuous certificate length, but also employed a decoupled framework and strict modality validation to precisely assess model performance across specific modalities. 
\section{VideoOdyssey}
\label{videoodyssey_construction}
\definecolor{lightorange}{rgb}{1.0, 0.9, 1.0}  
\definecolor{lightgreen}{rgb}{0., 0.66, 0.}    

\newcommand{\durationcolor}[1]{%
  \pgfmathsetmacro{\percent}{%
    (max(min(#1, 1000), 0)) / 1000 * 100
  }%
  \edef\temp{\noexpand\cellcolor{lightgreen!\percent!lightorange}}\temp#1%
}

\newcommand{\scorecolor}[1]{%
  \pgfmathparse{#1 < 5}
  \ifdim\pgfmathresult pt=1pt
    \cellcolor{lightorange}#1
  \else
    \pgfmathsetmacro{\percent}{%
      (max(min(#1, 110), 0) - 0) / (110 - 0) * 100
    }%
    \edef\temp{\noexpand\cellcolor{lightgreen!\percent!lightorange}}\temp#1%
  \fi
}

\begin{table*}[t]
\vspace{-0.5cm}
\centering
\caption{\textbf{Comparison of various benchmarks.} \textbf{Modality}: V and A denote video and audio. \textbf{Avg Len.}: average video duration (min). \textbf{Avg CCL.}: average continuous certificate length of questions. \textbf{Anno.}: A (automatic) or M (manual) annotation. \textbf{Multi-level CCL}: whether the benchmark covers multiple continuous certificate levels. \textbf{Open- domain}: whether the benchmark covers diverse video domains. \textbf{A-V Corr.}: whether answering questions requires audio-visual synchronization, and M indicates audio-visual, visual/audio-only questions are mixed. \textbf{Unimodal filter}: whether quality control strategies were used to exclude questions solvable by text-only or single-modality cues. \textit{See appendix \ref{appendix_e} for continuous certificate length estimation details.}
}
\vspace{-0.2cm}
\begin{adjustbox}{max width=1.0\textwidth}
\begin{tabular}{lccrrrrccccc}
\toprule
\textbf{Benchmarks} & \multicolumn{1}{c}{\textbf{Venue}} & \textbf{Modality}&\textbf{\#Videos}&\textbf{\makecell{Avg Len. \\ (min)}} & \textbf{\makecell{\#QA\\pairs}} &  \textbf{\makecell{Avg CCL \\ (min)}}&\textbf{Anno.} & \textbf{\makecell{Multi-level\\CCL}} & \textbf{\makecell{Open\\domain}}   & \textbf{\makecell{A-V\\Corr.}}  & \textbf{\makecell{Unimodal\\filter}} \\
\midrule
\multicolumn{12}{l}{\emph{\textbf{Long video benchmark}}}\\
MovieChat-1K~\citep{song2024moviechat} &CVPR'24& V & 130 &8.33 & 1,950 & 0.9 & M & \redxmark&  \redxmark   & -&  \redxmark\\
LongVideoBench~\citep{wu2024longvideobench}&NeurIPS'24 & V & 3,763 & 7.88 & 3,102 & 0.7 & M & \redxmark& \greencmark & -& \redxmark\\
Video-MMMU~\citep{hu2025video}&arXiv'25 & V & 300 & 8.44 & 900 & 3.6 & M & \redxmark& \redxmark  & - & \redxmark\\
MLVU~\citep{zhou2025mlvu} &CVPR'25 & V & 1,730 & 15.50 & 3,102 & 5.0 & A$\&$M& \redxmark & \greencmark  & -& \redxmark\\
Video-MME~\citep{fu2025video}&CVPR'25 & V & 900 & 16.97 & 2,700 & 6.0 & M & \redxmark& \greencmark &-  & \greencmark\\
CG-Bench~\citep{chen2024cg} &ICLR'25& V & 1,219 & 27.07 & 12,129 & 0.3  & M& \redxmark & \greencmark & - & \greencmark\\
InfiniBench~\citep{ataallah2025infinibench}&EMNLP'25 & V & 1,217 & 52.59 & 87,700 & 3.4 & A$\&$M& \redxmark & \redxmark  & -  & \greencmark\\
LVBench~\citep{wang2025lvbench}&ICCV'25 & V & 103 & 68.35 & 1,549 & 4.1 & M & \redxmark& \greencmark &- & \greencmark\\
\rowcolor{gray!10}
\textbf{VideoOdyssey-V} &-& V & 100 & \textbf{109.00} & 1,618 &\textbf{16.0} & M&\greencmark& \greencmark  & - & \greencmark\\

\midrule
\multicolumn{12}{l}{\emph{\textbf{Audio-Visual benchmark}}} \\
AVQA~\citep{yang2022avqa}&ACM MM'22 & V+A&57,000&0.17&57,335&0.1&M&\redxmark&\greencmark   &\greencmark &\redxmark \\
Music-AVQA~\citep{li2022learning} &CVPR'22& V+A & 9,288 & 1.00 & 45,867 & 0.4& M& \redxmark & \redxmark  & M & \redxmark\\
LongVALE~\citep{geng2025longvale}&CVPR'25 & V+A & 8,400 & 3.92 &- &0.4 & A$\&$M & \redxmark& \greencmark&M  & \redxmark \\
Daily-Omni~\citep{zhou2025daily} &arXiv'25&V+A&684&0.75&1,197&0.2&A$\&$M& \redxmark&\greencmark &\greencmark & \redxmark\\
OmniVideoBench~\citep{li2025omnivideobench} &arXiv'25& V+A & 628 & 6.40 & 1,000 &1.6 & M& \redxmark & \greencmark & \greencmark & \greencmark\\
LongInsightBench~\citep{han2025longinsightbench} &arXiv'25 &V+A & 1,001 & 8.99 & 4,781 & 0.9& A$\&$M& \redxmark  & \greencmark&\greencmark&\greencmark\\
WorldSense~\citep{hong2025worldsense} &ICLR'26& V+A & 1,662 & 2.35 & 3,172 &0.9  & M & \redxmark& \greencmark  &M  & \greencmark \\
LVOmniBench~\citep{tao2026lvomnibench} &arXiv'26& V+A & 275 & 34.50 & 1,014 & 0.8 & M & \redxmark& \greencmark  & \greencmark  & \greencmark\\
Video-MME-v2~\citep{fu2026video} &arXiv'26& V+A & 800 & 10.40 & 3,200 & 2.5 & M & \redxmark& \greencmark  &  M & \redxmark \\

\rowcolor{gray!10}
\textbf{VideoOdyssey-AV}&-&V+A&100&\textbf{109.00}&1,062&\textbf{12.8}&M&\greencmark&\greencmark&\greencmark&\greencmark \\
\bottomrule
\end{tabular}
\end{adjustbox}

\label{tab:dataset_comparison}
\vspace{-0.5cm}
\end{table*}

\begin{figure}[t] 
    \centering 
    \vspace{-0.1cm}
    \includegraphics[width=1.0\textwidth]{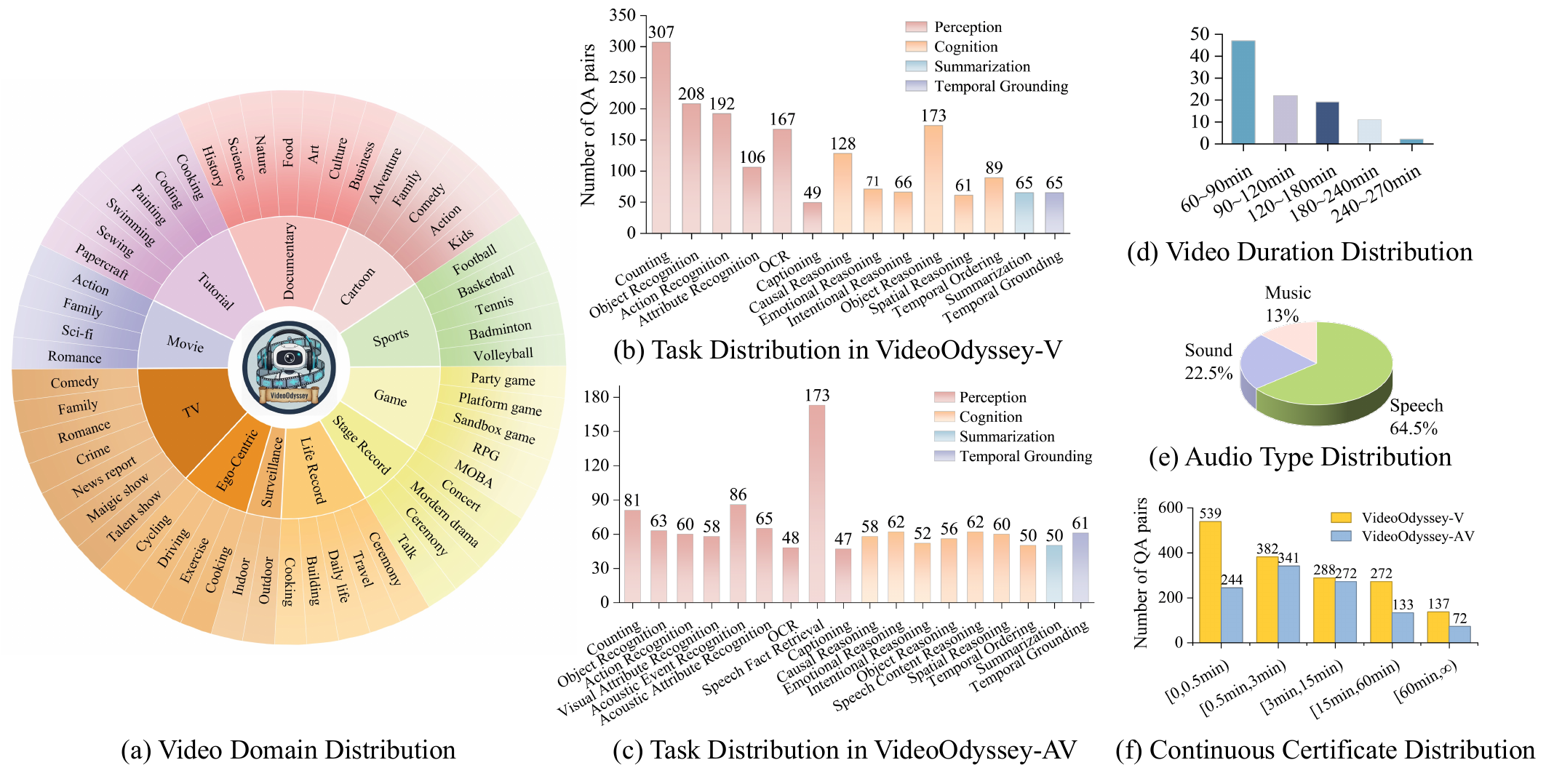} 
    \vspace{-0.6cm}
    \caption{Statistics of \textbf{\textit{VideoOdyssey}}. (a) \textit{VideoOdyssey} contains 11 domains and 54 subcategories. (b) \textit{VideoOdyssey-V} contains 1618 QA pairs across 14 tasks to assess model capability in four dimensions. (c) \textit{VideoOdyssey-AV} contains 1062 QA pairs across 18 tasks to asses model performance in four dimensions. (d) All videos exceed 60 minutes, with the longest over 4 hours. (e) \textit{VideoOdyssey-AV} features three audio types: speech, sound and music. (f) We design 5 granular continuous certificate length levels ranging from seconds to hours to precisely trace model performance.} 
    \label{fig:dataset_statistics} 
    \vspace{-0.4cm}
\end{figure}

In this section, we introduce \textbf{\textit{VideoOdyssey}}. Specifically, we detail the data collection in Sec. \ref{videoodyssey_data_collection}, the QA annotation process in Sec. \ref{videoodyssey_annotation}, and a rigorous quality control process in Sec. \ref{videoodyssey_quality_control}. The statistics of the dataset are summarized in Fig.~\ref{fig:dataset_statistics}.

\subsection{Data Collection}
\label{videoodyssey_data_collection}
To ensure the high quality and complexity of \textit{VideoOdyssey}, we adhere to a rigorous collection process centered on four principles: 1) All videos exceed 60 minutes to ensure sufficient temporal depth; 2) A minimum resolution of 720P is required for visual clarity; 3) Each video must contain dynamic scenes to provide substantial visual information; 4) Each video must contain rich audio to provide crucial auditory cues. Following these principles, we sourced 100 high-quality, synchronized audio-visual videos from YouTube. Accompanying subtitles were also downloaded if available, otherwise, we employed the Whisper-large-v3~\citep{radford2023robust} to generate subtitles for them. Ultimately, 89 videos are paired with subtitles, while 11 videos without distinct speech do not have subtitles.

\subsection{QA Annotation}
\label{videoodyssey_annotation}

\textit{Continuous certificate length} means the video length a human must continuously watch to definitively answer a given question. It explicitly quantifies the cognitive load imposed by continuous tracking, information integration, and memory retention across an ultra-long context.

During the QA annotation process, human annotators were permitted to freely navigate the video timeline and repeatedly review specific segments to design challenging multiple-choice questions. Crucially, they were required to label the precise continuous certificate length for each question. To guarantee the benchmark's difficulty and reliability, the annotation process strictly adhered to four core principles:
\textbf{1) Long-context dependency}: Annotators were required to annotate as many questions as possible that require sustained reasoning across long context.
\textbf{2) Modality dependency}: Questions in \textit{VideoOdyssey-V} must rely on visual cues, whereas questions in \textit{VideoOdyssey-AV} must necessitate the synergy of both audio and visual cues.
\textbf{3) Unambiguous clarity}: Questions must be objectively answerable without semantic ambiguity.
\textbf{4) Plausible distractors}: The three distractors must be semantically competitive and format-consistent with the ground-truth answer. 
Following these guidelines, human annotators designed 1,664 QAs for \textit{VideoOdyssey-V} and 1,141 QAs for \textit{VideoOdyssey-AV}.
These questions comprehensively evaluate model capability across four dimensions: perception, cognition, summarization, and temporal grounding (see Fig.~\ref{fig:dataset_statistics}(b-f)).

\subsection{Quality Control}
\label{videoodyssey_quality_control}
To ensure benchmark reliability, we implemented a rigorous two-stage quality control pipeline. First, an automated verification stage systematically eliminated evaluation shortcuts. For \textit{VideoOdyssey-V}, we used DeepSeek-R1 and GPT-4 to filter out any questions solvable via language priors alone. For \textit{VideoOdyssey-AV}, we used Gemini-2.5-Pro and Qwen3-Omni to discard any questions that could be answered using only video frames or only the audio track, thereby mandating true cross-modal synergy.
Following this automated filtering, a manual verification stage was conducted by human experts. They meticulously reviewed each remaining QA pair to confirm adherence to all annotation principles outlined in Sec~\ref{videoodyssey_annotation}, discarding any that failed. Through this dual-stage process, 46 questions were removed from \textit{VideoOdyssey-V} and 79 from \textit{VideoOdyssey-AV}, ultimately resulting in 1,618 and 1,062 high-quality QA pairs, respectively.

\section{Experiments}
\label{experiments}
\begin{table*}[t]
\centering
\footnotesize
\caption{\textbf{Performance of MLLMs on VideoOdyssey-V.} We show the performance of MLLMs on tasks across four dimensions. For both proprietary and open-source MLLMs, the highest and second-highest scores are highlighted in \textbf{bold} and \underline{underlined}, respectively.}
\label{tab:visual_main}
\vspace{-0.2cm}

\newcommand{\best}[1]{\textbf{#1}}
\newcommand{\second}[1]{\underline{#1}} 

\begin{adjustbox}{max width=1.0\textwidth}
\setlength{\tabcolsep}{2.5pt} 
\renewcommand{\arraystretch}{1.15}

\begin{tabular}{l c cccccc cccccc cc c}
\toprule
\multirow{2}{*}{\textbf{Model}} & \multirow{2}{*}{\shortstack{\# frms}}
& \multicolumn{6}{c}{\textbf{Perception}}
& \multicolumn{6}{c}{\textbf{Cognition}}
& \multicolumn{2}{c}{}  
& \multirow{2}{*}{\textbf{Overall}} \\
\cmidrule(lr){3-8} \cmidrule(lr){9-14} 

&  & Count & ObRec & AcRec & AtRec & OCR & Cap
& CaRea & EmRea & InRea & ObRea & SpRea & Order 
& Sum & TeGro
& \\
\midrule
\rowcolor{blue!5}
\multicolumn{17}{c}{\emph{\textbf{Human Baseline}}} \\
\midrule
Human & - &74.3&82.1&86.7&87.1&85.7&93.9&86.9&83.3&80.9&83.6&87.2&90.8&94.0&96.4&84.4\\

\midrule
\rowcolor{blue!5}
\multicolumn{17}{c}{\emph{\textbf{Proprietary MLLMs}}} \\
\midrule

GPT-5.2~\citep{singh2025openai} & 128
& 28.3 & 55.3 & 46.9 & 51.9 & \second{64.1} & 49.0 
& 51.6 & 43.7 & 60.6 & \second{60.1} & 45.9 & \second{69.7} 
& 44.6 & 44.6
& 49.0 \\

Gemini-2.5-Pro~\citep{comanici2025gemini} & 128
& 31.6 & \second{59.6} & 49.5 & 54.7 & 60.5 & 44.9 
& 52.3 & \best{49.3} & \second{62.1} & \second{60.1} & 41.0 & 64.0 
& \best{61.5} & 40.0
& \second{50.4} \\

Gemini-3.1-Pro~\citep{google2025gemini3} & 128
& \second{34.2} & \best{63.0} & \best{57.8} & \best{64.2} & \best{66.5} & \best{59.2} 
& \best{59.4} & \second{45.1} & \best{78.8} & \best{65.3} & \best{50.8} & \best{70.8} 
& \second{56.9} & \best{49.2}
& \best{56.3} \\

Seed-2.0-Pro\citep{seedseed2} & 128
& \best{34.9} & 51.9 & \second{54.7} & \second{61.3} & 50.9 & \second{57.1} 
& \second{57.8} & 43.7 & 57.6 & 52.6 & \second{47.5} & 56.2
& 55.4 & \second{47.7}
& 49.7 \\

\midrule

\rowcolor{blue!5}
\multicolumn{17}{c}{\emph{\textbf{Open-source Image LLMs}}} \\
\midrule

InternVL3.5-38B~\citep{wang2025internvl3} & 32
& \second{30.3} & 40.9 & 38.0 & 38.7 & 36.5 & 42.9 
& 42.2 & 39.4 & 36.4 & 35.8 & 29.5 & 36.0 
& 44.6 & 36.9
& 36.7 \\

Phi4-Multimodal~\citep{abouelenin2025phi} & 64
& 15.3 & 26.9 & 24.5 & 32.1 & 27.0 & 32.7 
& 32.8 & 31.0 & 28.8 & 27.2 & 37.7 & 14.6 
& 32.3 & 29.2
& 25.7 \\

Kimi-VL-A3B~\citep{team2025kimi} & 64
& 25.7 & 28.9 & 23.4 & 28.3 & 29.9 & 20.4 
& 28.9 & 26.8 & 24.2 & 24.3 & 21.3 & 22.5 
& 29.2 & 32.3
& 26.3 \\

LLaVA-Onevision-1.5-8B~\citep{an2025llava} & 128
& 21.5 & 37.5 & 25.0 & 33.0 & 32.3 & 32.7 
& 27.3 & 33.8 & 36.4 & 30.1 & 26.2 & 33.7 
& 29.2 & 35.4
& 29.6 \\

\midrule
\rowcolor{blue!5}
\multicolumn{17}{c}{\emph{\textbf{Open-source Video LLMs}}} \\
\midrule

Video-LLaVA-7B~\citep{lin2024video} & 8
& 14.7 & 20.7 & 20.8 & 29.3 & 19.2 & 36.7 
& 22.7 & 26.8 & 28.8 & 27.8 & 27.9 & 23.6 
& 23.1 & 18.5
& 22.3 \\

LLaVA-NeXT-Video-DPO-7B~\citep{zhang2024llavanextvideo} & 32
& 18.2 & 30.8 & 16.2 & 24.5 & 22.8 & 30.6
& 21.1 & 28.2 & 24.2 & 24.3 & 11.5 & 23.6
& 36.9 & 24.6
& 22.9 \\

LLaVA-NeXT-Video-DPO-34B~\citep{zhang2024llavanextvideo} & 32
& 16.6 & 26.4 & 26.6 & 34.0 & 27.5 & 34.7 
& 24.2 & 31.0 & 33.3 & 24.3 & 24.6 & 36.0 
& 30.8 & 26.2
& 25.7 \\

Video-R1-7B~\citep{feng2025video} & 64
& 24.4 & 36.5 & 33.3 & 34.0 & 38.3 & 36.7 
& 37.5 & 39.4 & 34.8 & 35.8 & 36.1 & 29.2 
& 36.9 & 36.9
& 33.7 \\

Video-KTR-7B~\citep{wang2026video} & 64
& 24.4 & 36.5 & 33.9 & 33.0 & 39.5 & 38.8 
& 41.4 & 36.6 & 37.9 & 35.3 & 31.1 & 23.6 
& 41.5 & 40.0
& 33.8 \\

VideoLLaMA3-7B~\citep{zhang2025videollama}& 64
& 28.7 & 28.4 & 29.2 & 37.7 & 29.9 & 14.3 
& 30.5 & 35.2 & 31.8 & 39.3 & 19.7 & 30.3 
& 24.6 & \second{43.1}
& 30.8 \\

Qwen3-VL-32B~\citep{bai2025qwen3} & 64
& 22.8 & \second{45.2} & 34.9 & 49.1 & 38.3 & 38.8 
& 34.4 & 33.8 & 40.9 & 43.4 & 36.1 & 40.5 
& 40.0 & 36.9
& 36.2 \\

Qwen3-VL-235B~\citep{bai2025qwen3} & 128
& 27.0 & 43.3 & \second{44.3} & 49.1 & 43.7 & \second{44.9} 
& 45.3 & 33.8 & 48.5 & 50.3 & 36.1 & 34.8 
& 47.7 & 38.5
& 40.6 \\

Qwen3.5-27B~\citep{team2026qwen3} & 128
& \best{30.9} & 43.8 & \best{50.0} & \best{59.4} & \second{48.5} & \best{51.0}
& \second{48.4} & \second{42.3} & \second{50.0} & \second{52.6} & \best{41.0} & \second{55.1}
& \second{49.2} & 21.5
& \second{44.6} \\

Kimi-K2.5~\citep{team2026kimi} & 128
& 26.7 & \best{57.7} & \best{50.0} & \second{51.9} & \best{52.7} & \second{44.9} 
& \best{57.0} & \best{52.1} & \best{62.1} & \best{53.8} & \second{39.3} & \best{57.3} 
& \best{69.2} & \best{46.2}
& \best{48.6} \\

\bottomrule
\end{tabular}

\end{adjustbox}
\vspace{-0.2cm}
\end{table*}
\begin{table*}[t]
\centering
\footnotesize
\caption{\textbf{Performance of MLLMs on VideoOdyssey-AV.} We show the performance of MLLMs on tasks across four dimensions. For both proprietary and open-source MLLMs, the highest and second-highest scores are highlighted in \textbf{bold} and \underline{underlined}, respectively.}
\label{tab:av_main}

\vspace{-0.2cm}
\newcommand{\best}[1]{\textbf{#1}}
\newcommand{\second}[1]{\underline{#1}} 

\begin{adjustbox}{max width=1.0\textwidth}
\setlength{\tabcolsep}{2.5pt}

\renewcommand{\arraystretch}{1.15}

\begin{tabular}{l c ccccccccc ccccccc cc c}
\toprule
\multirow{2}{*}{\textbf{Model}} & \multirow{2}{*}{\shortstack{\# frms}}
& \multicolumn{9}{c}{\textbf{Perception}}
& \multicolumn{7}{c}{\textbf{Cognition}}
& \multicolumn{2}{c}{} 
& \multirow{2}{*}{\textbf{Overall}} \\
\cmidrule(lr){3-11} \cmidrule(lr){12-18} 

&  & Count & ObRec & AcRec & VAR & AER & AAR & OCR & SFR & Cap
& CaRea & EmRea & InRea & ObRea & SCR & SpRea & Order 
& Sum & TeGro
& \\
\midrule
\rowcolor{blue!5}
\multicolumn{21}{c}{\emph{\textbf{Human Baseline}}} \\
\midrule
Human &-&75.0&85.0&85.4&81.0&78.7&71.1&81.1&87.9&80.6&81.0&74.4&79.3&73.7&75.0&82.6&71.0&70.8&93.2&80.7\\
\midrule
\rowcolor{blue!5}
\multicolumn{21}{c}{\emph{\textbf{Proprietary Omni-Modal LLMs}}} \\
\midrule

Gemini-2.5-Pro~\citep{comanici2025gemini} & 64
& 25.9 & 41.3 & \second{40.0} & \second{44.8} & \best{38.4} & 44.6 & \second{45.8} & 50.3 & \best{74.5}
& 50.0 & 46.8 & \second{48.1}& \best{42.9} & \best{53.2} & 23.3 & \second{38.0} & 60.0 & 37.7 & 43.9 \\

Gemini-3-Flash~\citep{google2025gemini3} & 64
& \second{30.9} &44.4 & \best{45.0} & 36.2 & 32.6 & \second{46.2} & \second{45.8} & \second{50.9} & 66.0
& \best{56.9} & \best{51.6} & \second{48.1} & \best{42.9} & \best{53.2} & \second{30.0} & \best{44.0} & 50.0 & 32.8
& \second{44.3} \\

Gemini-3.1-Pro~\citep{google2025gemini3} & 64
& 28.4 & \best{52.4} & \second{40.0} & 43.1 & \second{37.2} &41.5 & \best{52.1} & \best{59.5} & 61.7
& \second{53.5} & 46.8 & \best{65.4} & \second{41.1}&41.9 & 28.3 &34.0 & \best{64.0} &\second{39.3}
& \best{46.1} \\

Qwen3.5-Omni-Plus~\citep{qwenteam2026qwen35omnitechnicalreport} & 64
& \best{37.0} & \second{50.8} & 38.3 & \best{46.6} & \best{38.4} &\best{52.3} & 33.3 & 43.4 & \second{72.3}
& 50.0 & \second{48.4} & 40.4 & 35.7& \second{45.2} & \best{31.7} &20.0 & \second{62.0} &\best{42.6}
& 43.0 \\
\midrule
\rowcolor{blue!5}
\multicolumn{21}{c}{\emph{\textbf{Open-source Omni-Modal LLMs}}} \\
\midrule

OneLLM-7B~\citep{han2024onellm} & 64
& 12.4 & 27.0 & 21.4 & 22.4 & 19.8 & 21.5 & \best{33.3} & 21.4 & 21.3
& 17.2 & 30.7 & 19.2 & 21.4 & 24.2 & 23.3 & 16.0 & 24.0 & 24.6
& 21.2 \\

VideoLLaMA2-7B~\citep{cheng2024videollama} & 64
& 16.1 & 22.2 & 16.6 & \second{25.9} & 20.9 & \second{32.3} & 25.0 & 20.8 & \best{46.8}
& 17.2 & \second{33.9} & 9.62 & \best{30.4} & 25.8 & 23.3 & 26.0 & 32.0 & 31.2
& 24.1 \\

Unified-IO-2 L~\citep{lu2024unified} & 64
& 13.6 & 28.6 & 21.7 & \best{27.6} & 25.6 & 21.5 & 22.9 & 26.0 & 36.2
& 22.4 & 27.4 & 25.0 & 19.6 & 24.2 & 25.0 & \best{32.0} & 24.0 & 19.7
& 25.1 \\

Unified-IO-2 XL~\citep{lu2024unified} & 64
& \second{28.4} & 22.2 & 28.3 & 24.1 & 24.4 & 29.2 & 27.1 & 23.1 & 36.2
& 20.7 & 29.0 & \second{26.9} & 21.4 & 25.8 & 23.3 & \second{24.0} & 28.0 & 21.3
& 26.4 \\

Unified-IO-2 XXL~\citep{lu2024unified} & 64
& 23.5 & 25.4 & 23.3 & \best{27.6} & \second{27.9} & \second{32.3} & 20.8 & 22.0 & \second{44.7}
& 22.4 & 22.6 & 23.1 & \second{28.6} & 24.2 & \best{28.3} & 20.0 & 20.0 & 31.2
& 25.9 \\

Ola-7B~\cite{liu2025ola} & 64
& \best{29.6} & \second{30.2} & 23.3 & \best{27.6} & \best{30.2} & \best{36.9} & 25.0 & \second{28.3} & 42.6
& \best{31.0} & \best{35.5} & \best{38.5} & \best{30.4} & \best{43.6} & \second{26.7} & 20.0 & \best{36.0} & \second{32.8}
& \best{31.5} \\

Qwen3-Omni-30B~\citep{xu2025qwen3} & 64
& 24.7 & \best{34.9} & \best{36.7} & 22.4 & 25.6 & 29.2 & \second{29.2} & \best{28.9} & 38.3
& \second{29.3} & 29.0 & \second{26.9} & 23.2 & \second{29.0} & 16.7 & 18.0 & \second{34.0} & \best{37.7}
& \second{28.7} \\

VITA-1.5-7B~\citep{fu2025vita} & 16
& 23.5 & 28.6 & \second{30.0} & 20.7 & 20.9 & 29.2 & 27.1 & 22.5 & 6.4
& 24.1 & 29.0 & 11.5 & 30.4 & 30.7 & 16.7 & 20.0 & 14.0 & 13.1
& 22.6 \\

\bottomrule
\end{tabular}
\end{adjustbox}
\vspace{-0.6cm}
\end{table*}
\subsection{Settings}
To comprehensively evaluate the performance of current MLLMs, we conducted extensive experiments on the \textit{VideoOdyssey-V} and \textit{VideoOdyssey-AV} benchmarks. On \textit{VideoOdyssey-V}, we assessed a total of 18 MLLMs, including proprietary MLLMs (\eg, GPT-5.2, Gemini-3.1-Pro), open-source image LLMs (\eg, InternVL3.5, LLaVA-Onevision-1.5) and open-source video LLMs (\eg, Qwen3.5, Kimi-K2.5). For \textit{VideoOdyssey-AV}, we evaluate 12 omni-modal LLMs, including proprietary omni-modal LLMs (\eg, Gemini-3.1-Pro, Qwen3.5-Omni-Plus) and open-source omni-modal LLMs (\eg, Ola, Qwen3-Omni). On \textit{VideoOdyssey-V}, proprietary MLLMs utilize 128 frames, while open-source models use their maximum configurations. For \textit{VideoOdyssey-AV}, a 64-frame sampling is generally applied—with VITA-1.5-7B utilizing 16 frames—to ensure temporal coverage for ultra-long tasks where default sparse sampling often fails. All model outputs were evaluated through direct comparison with ground-truth answers.
\begin{figure}[t] 
    \centering 
    \includegraphics[width=1.0\textwidth]{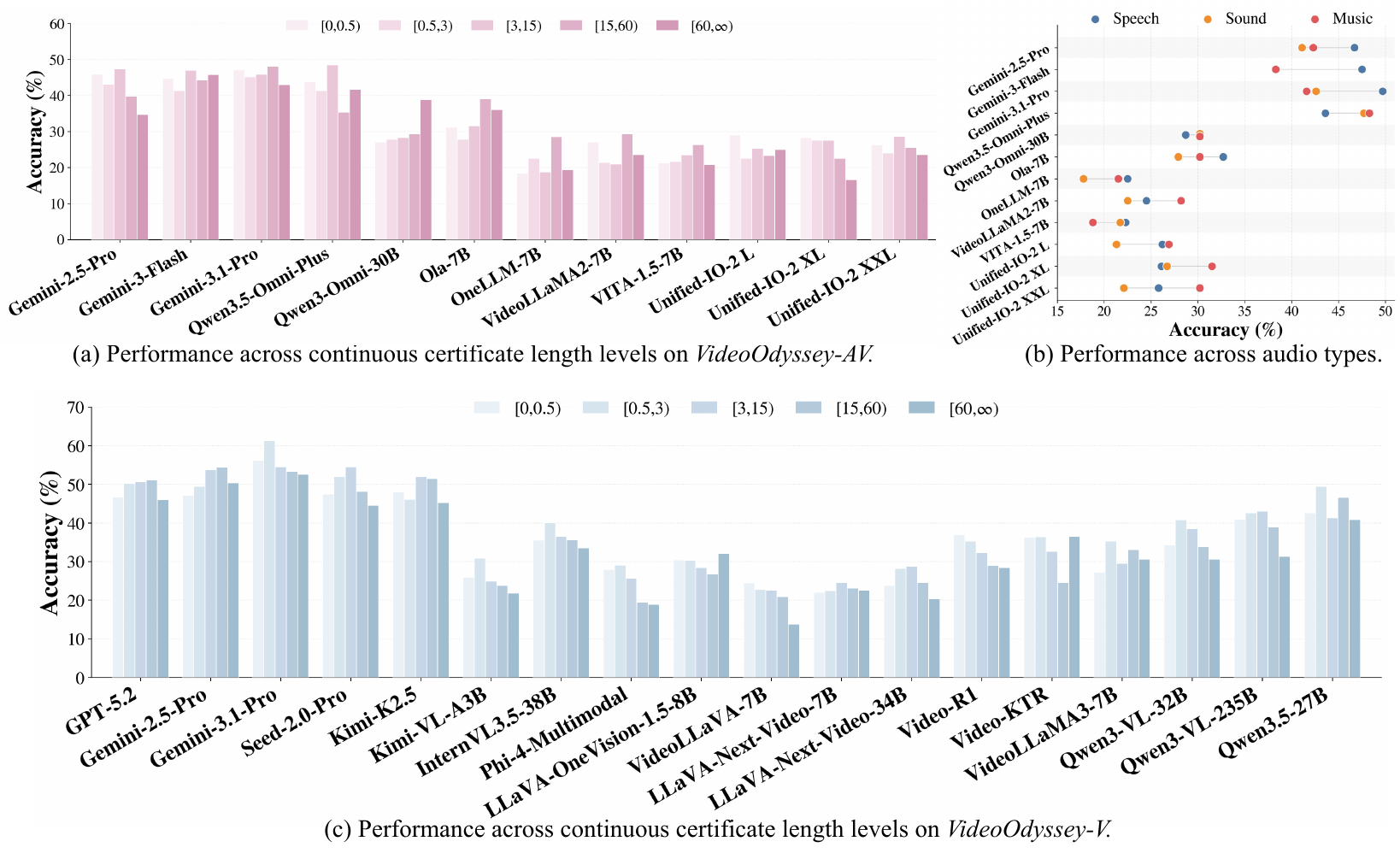} 
    \vspace{-0.5cm}
    \caption{Performance of MLLMs across five continuous certificate length levels on \textit{VideoOdyssey-AV} {(a)} and \textit{VideoOdyssey-V} {(c)}, and across three audio types on \textit{VideoOdyssey-AV} {(b)}. \textit{See appendix \ref{appendix_d} for more results.}}
    \label{fig:performance_across_CCL_and_audio_types}
    \vspace{-0.4cm}
\end{figure}

\vspace{-0.1cm}

\subsection{Main Results and Findings}
\label{experiments:main}

Table \ref{tab:visual_main} and \ref{tab:av_main} report the main benchmarking results. Fig. \ref{fig:performance_across_CCL_and_audio_types} shows the investigation on the continuous certificate length and audio types. The key observations are discussed below.

\noindent \textbf{\textit{Current MLLMs struggle with ultra-long-context and omni-modal settings.}}
Gemini-3.1-Pro leads \textit{VideoOdyssey-V} with only 56.3\%, barely reaching a passing threshold. The challenge is even more pronounced on \textit{VideoOdyssey-AV}, where Gemini-3.1-pro only scored 46.1\%. 
Human evaluators achieve 84.4\% and 80.7\% on \textit{VideoOdyssey-V} and \textit{VideoOdyssey-AV}, respectively, highlighting a massive gap between current models and human cognition.
Task-specific analysis reveals that counting remains an significant challenge across both settings. On \textit{VideoOdyssey-V}, GPT-5.2 achieves only 28.3\%, with difficulty intensifying on \textit{VideoOdyssey-AV} where Gemini-3.1-Pro achieves only 28.4\%. Notably, Seed-2.0-Pro stands out in visual perception, topping counting at 34.9\% and ranking second behind Gemini-3.1-Pro across many perception tasks. Qwen3.5-Omni-Plus exhibits superior audio-visual perception, securing a landslide counting lead (37.0\%) alongside strong performance in other perception tasks. Despite these highlights, models universally struggle with spatial reasoning and temporal grounding in both settings. Furthermore, dismal scores on acoustic event recognition expose a critical deficiency in non-verbal audio comprehension.

\noindent \textbf{\textit{A major gap remains between open-source and proprietary models.}}
On \textit{VideoOdyssey-V}, the leading open-source model, Kimi-K2.5 (48.6\%), lags behind Gemini-3.1-Pro by 7.7\%. This gap persists on \textit{VideoOdyssey-AV}, where the leading open-source model, Ola-7B (31.5\%) trails Gemini-3.1-Pro by 14.6\%; notably, the majority of other open-source models merely hover around the random-guess baseline. This gap reflects deficient native cross-modal fusion in open-source models. Unlike unified proprietary architectures, most open-source models rely on modular ``plug-and-play'' designs that struggle with the high information density of long audio-visual streams. Consequently, extra modalities often act as noise, interfering with rather than aiding the reasoning process.


\noindent \textbf{\textit{Long continuous certificate length presents significant challenges.}}
As shown in Fig. \ref{fig:performance_across_CCL_and_audio_types}, most models peak at short continuous certificate (under 3 mins) but degrade significantly as it increases, struggling to capture long-range dependencies due to memory constraints. GPT-5.2 and Gemini-2.5-Pro do not exhibit such severe decay. Notably, the accuracy of Gemini-2.5-Pro progressively improves as the certificate length expands, dropping only at the extreme [60, $\infty$) interval. 
Conversely, the audio-visual setting exhibits fluctuating and inconsistent performance across continuous certificate lengths. This likely stems from the relative immaturity of audio-visual integration compared to pure vision. The simultaneous demands of localization, cross-modal alignment, and long-context reasoning induce severe cognitive overload, causing high variance that obscures true model capabilities. 

\noindent \textbf{\textit{Performance varies across audio types.}}
As shown in Fig. \ref{fig:performance_across_CCL_and_audio_types}, the Gemini family exhibits a strong ASR (Automatic Speech Recognition) bias, relying heavily on verbal speech cues while struggling with non-verbal acoustics (sound and music). Broadly, environmental sound comprehension remains a severe weakness across most models.
Paradoxically, many open-source models perform worse on speech than on music. We hypothesize this anomaly arises because speech tasks inherently require ultra-long context tracking, which is a critical bottleneck of these models. Notably, the Qwen-Omni series emerges as a strong exception. Both its open and proprietary models demonstrate exceptional non-verbal understanding, with Qwen3.5-Omni-Plus achieving the strongest comprehension capabilities in both sound and music tasks.

\begin{figure}[t] 
    \centering 
    \vspace{-0.2cm}
    \includegraphics[width=1.0\textwidth]{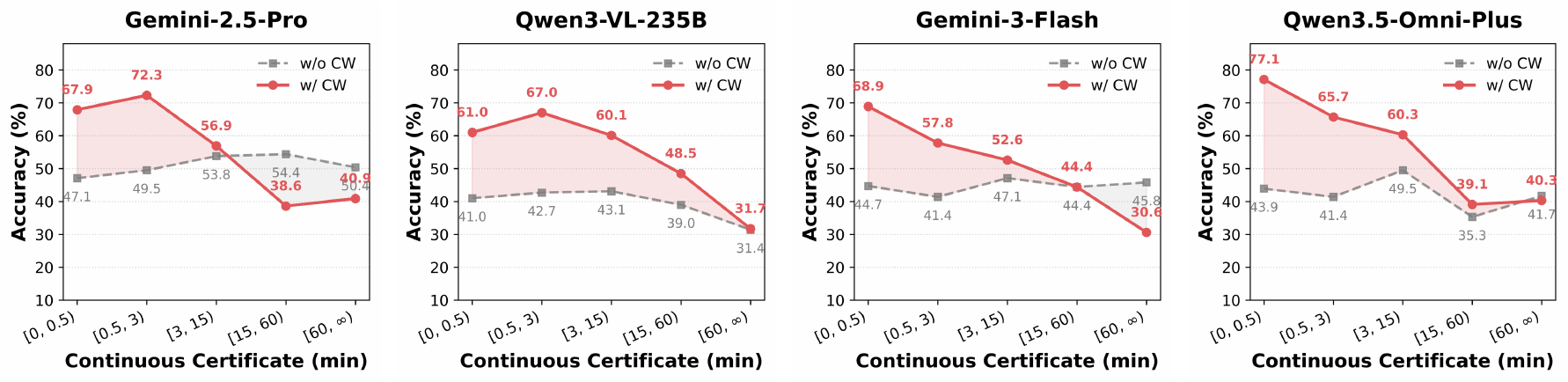} 
    \vspace{-0.5cm}
    \caption{Impact of certificate window (CW) on selected models across different continuous certificate length levels. We show the performance of Gemini-2.5-Pro and Qwen3-VL-235B on \textit{VideoOdyssey-V} and the performance of Gemini-3-Flash and Qwen3.5-Omni-Plus on \textit{VideoOdyssey-AV}.}
    \label{fig:impact_of_CW}
    \vspace{-0.5cm}
\end{figure}

\subsection{Further Analysis}
\label{experiments_discussion}

To isolate retrieval deficits from fundamental reasoning flaws, we evaluate models with ground-truth certificate windows (CW) across varying continuous certificate lengths and input modalities. Subsequently, we investigate if retrieval-based agentic method can mitigate these bottlenecks.

\noindent \textbf{\textit{How do models behave when ground-truth certificate windows are directly provided?}}
Fig. \ref{fig:impact_of_CW} reveals that the impact of certificate windows is highly dependent on the continuous certificate length.
This exposes critical insights into the core failure modes of current MLLMs.

\noindent \textbf{\textit{1) Gains on short clips reveal a search bottleneck, yet expose fundamental comprehension flaws.}}
Providing certificate windows yields dramatic gains for shorter clips (< 3 minutes), confirming a severe ``needle-in-a-haystack'' retrieval deficit. For instance, accuracy on the [0, 0.5) interval consistently jumps by over 20\%. However, absolute performance remains unsatisfactory. Notably, even with the exact clips provided, accuracy on the [0, 0.5) interval is frequently below that of the [0.5, 3) interval on \textit{VideoOdyssey-V}. This indicates that beyond mere retrieval difficulties, models face a specific bottleneck in fine-grained perception.
Furthermore, on \textit{VideoOdyssey-AV}, accuracy on the [0.5, 3) interval remains surprisingly low. For example, Gemini-3-Flash fails to reach a passing grade (57.8\%). This demonstrates that while the retrieval bottleneck is a significant hurdle, models' foundational ability to reason over dense audio-visual cues is a more critical limitation.

\noindent \textbf{\textit{2) Performance degradation and inversion on long clips due to information density.}}
As continuous certificate length increases, model performance steadily declines (beyond 3 minutes on \textit{VideoOdyssey-V} and 0.5 minutes on \textit{VideoOdyssey-AV}). Strikingly, when the continuous certificate exceeds 15 minutes, accuracy often drops to baseline levels or even lower. This highlights the challenge of high information density. Processing these long segments induces severe cognitive overload, preventing models from maintaining unbroken logical chains. In such cases, the redundant tokens in a dense clip are more disruptive than scanning the entire video.

\begin{figure}[t] 
    \centering 
    \includegraphics[width=1.0\textwidth]{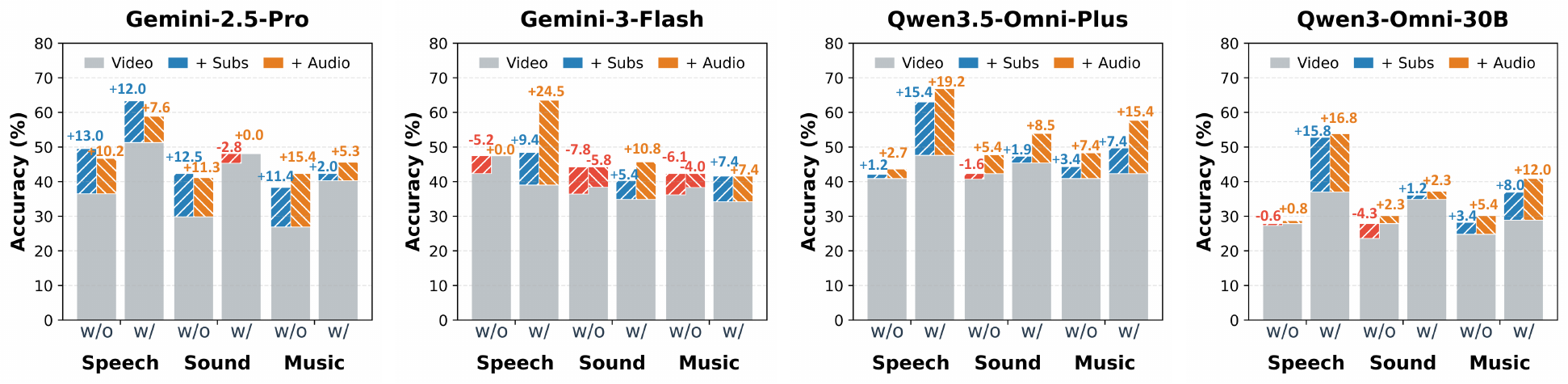} 
    \vspace{-0.7cm}
    \caption{Impact of different inputs for selected models across three audio types on \textit{VideoOdyssey-AV}, under w/o CW and w/ CW settings. Red values indicate performance drops.}
    \label{fig:impact_of_diff_inputs_w_wo_CW}
    \vspace{-0.4cm}
\end{figure}

\noindent \textbf{\textit{How do input modalities impact models when ground-truth certificate windows are directly provided?}} 
Fig. \ref{fig:impact_of_diff_inputs_w_wo_CW} reveals that removing the retrieval burden dramatically alters the benefits of additional modalities, exposing fundamental cross-modal bottlenecks and semantic biases in current MLLMs.

\noindent \textbf{\textit{1) Divergent marginal returns in proprietary models.}}
Providing the ground truth certificate window fundamentally alters the marginal benefits of additional modalities. For Gemini-2.5-Pro, gains from adding subtitles or audio shrink noticeably compared to the full-video setting. This suggests a ceiling effect: the isolated visual clip provides sufficient context, leading to diminishing returns or even negative interference. Conversely, Gemini-3-Flash experiences a massive rebound, transforming the severe performance drops observed in full-video settings into substantial gains. Mechanistically, this highlights how temporal constraints dictate fusion efficiency: Pro's high-capacity architecture rapidly saturates on localized visual features, rendering extra modalities redundant. In contrast, Flash's lightweight architecture exhibits strict alignment sensitivity; while additional modalities act as distractors in full videos, precisely temporal grounding unlock its multi-modal synergy.

\noindent \textbf{\textit{2) Disproportionate gains skewed towards speech tasks.}}
Despite overall improvements, modality-driven gains remain heavily skewed. Across most models, performance leaps on speech tasks vastly outpace those on sound and music. Strikingly, this trend even holds for Qwen3.5-Omni-Plus and Qwen3-Omni-30B—models that originally favored non-verbal tasks in full-video settings but exhibit a drastic reversal once ground-truth certificate window is provided. This exposes a deep-rooted semantic bottleneck: even in ideal short-context scenarios, current architectures are predominantly optimized for text-like spoken dialogue. Deeply fusing and reasoning over non-verbal acoustic semantics remains a significant challenge.

\begin{wraptable}{r}{0.5\textwidth}
    \vspace{-1.5em}
    \centering
    \setlength{\tabcolsep}{3pt} 
    \caption{Performance of Deep Video Discovery.}
    \vspace{-0.2cm}
    \label{tab:dvd}
    \begin{adjustbox}{max width=\linewidth}
    \begin{tabular}{lcccccc}
        \toprule
        \textbf{Model} & [0,0.5) & [0.5,3) & [3,15) & [15,60)& [60,$\infty$) &Overall\\
        \midrule
        GPT-4.1-mini & 43.8 &29.7 &60.0  & 40.0 & 35.0 & 40.7\\
        \rowcolor{gray!20}
        DVD (GPT-4.1-mini) & 45.8 & 51.4 & 40.0 & 25.0 & 30.0 & 41.3 \\
        \midrule
        GPT-5.2 & 56.3 & 48.7 & 56.0 & 50.0 & 35.0& 50.7 \\
        \rowcolor{gray!20}
        DVD (GPT-5.2) & 64.6 & 51.4& 48.0 & 45.0 & 40.0 & 52.7 \\
        \bottomrule
    \end{tabular}
    \end{adjustbox}
    \vspace{-1em}
\end{wraptable}
\noindent \textbf{\textit{How does the retrieval-based agentic method perform?}} 
We evaluate the Deep Video Discovery (DVD)~\citep{zhang2025deep} on a representative subset of 11 videos (one from each domain) and 150 QA pairs. Using GPT-4.1~\citep{openai2025gpt4.1} for captioning and o4-mini~\citep{openai2025o3} for reasoning, we specifically alternate the frame inspection VLM, testing both GPT-5.2 and GPT-4.1-mini. Table~\ref{tab:dvd} shows that DVD offers only a marginal improvement over the base model. Crucially, the improvement is predominantly concentrated in short-span questions while performance on longer continuous certificate lengths degrades compared to the base model. The discrepancy can be attributed to the inherent design of the retrieval-based pipeline: search-based agents are primarily effective at pinpointing localized evidence for short-span queries while weak at coping with long-term logical chains.
\section{Conclusion}
\textit{VideoOdyssey} is a comprehensive benchmark for evaluating MLLMs in authentic ultra-long video scenarios. Thanks to the continuous certificate length metric, \textit{VideoOdyssey} exposes staggering  gaps in state-of-the-art models and reveals fundamental comprehension bottlenecks rather than simple search failures: current models struggle with fine-grained perception in short spans and consistently fail to maintain long-term logical chains across massive time spans. We hope \textit{VideoOdyssey} will drive the evolution of MLLMs toward genuine real-world video understanding.
\section{Acknowledgment}
This research is supported by Hong Kong Research Grants Council Early Career Scheme (No. 22200824), General Research Fund (No. 12200526), and HKBU Rising Star Research Grant.






\bibliographystyle{plainnat}
\bibliography{ref}

\clearpage
\appendix

\section{Technical appendices and supplementary material}

Construction of contents:
\begin{itemize}
    \item \ref{appendix_b}: Definition and example of each task type
    \item \ref{appendix_c}: More statistics of our dataset
    \item \ref{appendix_d}: More results
    \item \ref{appendix_e}: Details for evaluating certificate lengths 
    \item \ref{appendix_f}: Details for evaluation with ground-truth certificate window
    \item \ref{appendix_g}: Evaluation prompts
    \item \ref{appendix_h}: Failure case study
    \item \ref{appendix_i}: Limitations and broader impacts
\end{itemize}

\section{Definition and example of each task type}
\label{appendix_b}

Table \ref{tab:v_task_types} and Table \ref{tab:av_task_types} show the definition and an example of each task in \textit{VideoOdyssey-V} and \textit{VideoOdyssey-AV}.

\vspace{-0.4cm}
\begin{longtable}{p{2cm} | p{2cm} | p{3.5cm} | p{6.5cm}}
\caption{Details of task types in \textit{VideoOdyssey-V}} \label{tab:v_task_types}\\
\hline
\textbf{Level} & \textbf{Task Type} & \textbf{Definition} & \textbf{Example} \\ 
\hline
\endfirsthead
\hline
\textbf{Level} & \textbf{Task Type} & \textbf{Definition} & \textbf{Example} \\ 
\hline
\endhead

\hline
\multicolumn{4}{r}{\textit{Continued on next page...}} \\
\endfoot      

\hline
\endlastfoot   

\textbf{Perception}
 & Counting 
 & Count the occurrences of specific entities in the video, including humans, other objects, scenes and events. 
 & How many lattes with oat milk were sold in the first hour of the video? \newline Options: \newline A. 6 \newline \textcolor{green!50!black}{B. 5} \newline C. 3 \newline D. 4  \\
\cline{2-4} 

 & Object \newline Recognition 
 & Recognize and classify specific objects presented in the video.
 & Which of the following modes of transportation is never seen in the video?
 \newline Options: \newline A. Bike \newline B. Train \newline C. Ship \newline \textcolor{green!50!black}{D. Truck} \\ 
\cline{2-4} 

& Action \newline Recognition & Identify the actions of humans or other objects in the video.
 & What did the boy wearing a white hoodie and jeans do when he left his seat for the second time?
 \newline Options: \newline A. Hand in materials to the teacher
 \newline B. Leave the classroom \newline C. Go to the podium to interact
 \newline \textcolor{green!50!black}{D. Throw away trash
} \\ 
\cline{2-4} 

& Attribute \newline Recognition & Identify the specific visual attributes of entities, including humans, objects and scenes.
 & What type of top is actor Reggie Rocc wearing in his third costume for this stage play?
 \newline Options: \newline \textcolor{green!50!black}{A. Causal shirt}
 \newline B. Hoodie \newline C. Sweater
 \newline D. Suit jacket \\ 
\cline{2-4} 

& OCR & Recognize textual information that appears in the video.
 & Who was the first-place finisher in the sixth swimming race?
 \newline Options: \newline A. Luke Hobson
 \newline \textcolor{green!50!black}{B. Brendan Burns} \newline C. Liam Bell
 \newline D. Josh Liendo
 \\ 
\hline

 \textbf{Perception}& Captioning & Generate a text description that details the specific visual actions, objects, and scene dynamics observable in the video.
 & Describe the scene where a digital HUD appears on screen with the caption: 'THEY AIM TO DESTROY AND DECEIVE’.
 \newline Options: \newline \textcolor{green!50!black}{A. A robot stands with its back to the camera, swaying its body, facing a large group of identical mechanical soldiers arranged in neat formation.}
 \newline B. A robot sways its body as it walks slowly, holding a gun in its hand, with a group of robotic soldiers arranged in neat formation in the background.
 \newline C. A scene set in outer space, rendered in orange tones, shows visible planets and floating asteroids.
 \newline D. A group of astronauts and a group of robots engage in a battle at a base on an extraterrestrial planet.
 \\ 
\hline

\textbf{Cognition} 
 & Causal \newline Reasoning
    & Infer the underlying causes or resulting consequences of a specific event.
    & What is the reason the pig grew elephant ears and a trunk? \newline Options: \newline A. It ate the two wolves' poison. \newline B. It was enchanted by the magical girl. \newline \textcolor{green!50!black}{C. It ate poisonous wolfberries.} \newline D. It was bitten by a poisonous mosquito.\\
\cline{2-4}

& Emotional \newline Reasoning
    & Infer the emotional state, underlying causes, and evolutionary trajectories of specific entities or the overall atmosphere.
    & How does the daughter's emotion towards her father change in the video? \newline A. Disappointment -> Anger -> Gratitude -> Unforgettable. \newline B. Estrangement -> Confusion -> Understanding -> Admiration \newline C. Pity -> Disappointment -> Resentment -> Sympathy. \newline \textcolor{green!50!black}{D. Resentment -> Dependence -> Sympathy -> Dependence.}\\
\cline{2-4}

& Intentional \newline Reasoning
    & Infer the underlying purposes or motivations behind a specific character’s actions.
    & What is the primary goal of the two wolves kidnapping the girl? \newline A. To get revenge on the bear for hitting them earlier. \newline B. To make the girl do housework for them. \newline \textcolor{green!50!black}{C. To demand food in the refrigerator from the bear.} \newline D. To make the bear respect them more.\\
\cline{2-4}

& Object \newline Reasoning
    & Infer a specific object that meets a certain condition, or infer its function, attributes, or relationships between objects.
    & Based on their behavior during class, who do you think was the least engaged among the girl in the light purple top, the girl in the pink top, the boy in the green top, and the boy in the white hoodie? \newline A. The girl in the light purple top \newline B. The girl in the pink top \newline C. The boy in the green top \newline \textcolor{green!50!black}{D. The boy in the white hoodie}\\
\hline

\textbf{Cognition}& Temporal \newline Ordering
    & Arrange multiple key visual events from the video in temporal order.
    & Please arrange the following students in chronological order based on the time of their last appearance in the video: 1. the girl in the purple top, 2. the boy wearing khaki shorts, 3. the boy in the white hoodie, 4. the girl in the pink top. \newline \textcolor{green!50!black}{A. 1432} \newline B. 1342 \newline C. 1234 \newline D. 1243\\
\cline{2-4}

& Spatial \newline Reasoning
    & Reason about the spatial locations of objects, including relative directions and the paths between them.
    & If I stand next to the cabinet for bowls and chopsticks in the kitchen, facing the cabinet, is the dining table to my front-left, front-right, back-left, or back-right?"\newline A. Front-left \newline B. Front-right \newline \textcolor{green!50!black}{C. Back-left} \newline D. Back-right\\
\hline

\textbf{Summarization}& Summarization
    & Analyze the visual information to achieve a high-level, abstract understanding.
    & Please summarize Swiatek's performance in the sixth game of the first set.\newline A. Swiatek was down 0-40, kept fighting back, got Advantage first, and won the game. \newline B. Swiatek was down 0-30, kept fighting back, got Advantage first, and won the game. \newline C. Swiatek was down 0-40, kept fighting back, and won the game after the opponent had Advantage. \newline \textcolor{green!50!black}{D. Swiatek was down 0-30, kept fighting back, and won the game after the opponent had Advantage.}\\
\hline

\textbf{Temporal \newline Grounding}& Temporal \newline Grounding
    & Locate the specific timestamp in the video where a described visual event occurs.
    & What is the exact timestamp when the flamingo first and last appears in the video?\newline A. 00:02:21-02:16:40 \newline B. 00:02:22-02:16:30 \newline C. 00:02:23-02:16:00 \newline \textcolor{green!50!black}{D. 00:02:24-02:16:42}\\
\hline
\end{longtable}
\newpage
\begin{longtable}{p{2cm} | p{2cm} | p{3.5cm} | p{6.5cm}}
\caption{Details of task types in \textit{VideoOdyssey-AV}} \label{tab:av_task_types}\\
\hline
\textbf{Level} & \textbf{Task Type} & \textbf{Definition} & \textbf{Example} \\ 
\hline
\endfirsthead
\hline
\textbf{Level} & \textbf{Task Type} & \textbf{Definition} & \textbf{Example} \\ 
\hline
\endhead

\hline
\multicolumn{4}{r}{\textit{Continued on next page...}} \\
\endfoot      

\hline
\endlastfoot   

\textbf{Perception}
 & Counting 
 & Count the occurrences of specific entities in the video based on both visual and audio cues. 
 & During the first ten minutes of class, how many times did Abby participate in discussions?  \newline Options: \newline A. 1 \newline B. 2 \newline C. 3 \newline \textcolor{green!50!black}{D. 4}  \\
\cline{2-4} 

 & Object \newline Recognition 
 & Recognize and classify specific objects presented in the video, based on both visual and audio cues.
 & Who is the student who went to the front of the classroom to interact with the teacher?
 \newline Options: \newline A. Eric \newline B. Bernadette \newline \textcolor{green!50!black}{C. Abby} \newline D. Dylan \\ 
\cline{2-4} 

 & Action \newline Recognition 
 & Identify the actions of humans or other objects in the video, based on both visual and audio cues.
 & What did Abby do the first time she left her seat during class?
 \newline Options: \newline A. Hand in materials to the teacher \newline B. Leave the classroom \newline \textcolor{green!50!black}{C. Go to the podium to interact} \newline D. Throw away trash \\ 
\cline{2-4}

 & Visual \newline Attribute \newline Recognition 
 & Identify the visual attributes of specific entities, including humans, objects and scenes, based on both visual and audio cues. 
 & What color clothing was the person wearing who was the main focus of the conversation the second time money was mentioned in the video?  \newline Options: \newline A. Black \newline B. Green \newline C. Brown \newline \textcolor{green!50!black}{D. Red}  \\
\cline{2-4}

 & OCR 
 & Recognize textual information that appears in the video, based on both visual and audio cues. 
 & On day 1 in Istanbul, what is the name of the restaurant that locals consider the king of breakfast?  \newline Options: \newline A. Fehmi Pasha Café \newline B. Karaköy Breakfast House \newline C. Galata Traditional Kitchen \newline \textcolor{green!50!black}{D. Hasan Fehmi Özsüt Karaköy Muhallebicisi}  \\
\cline{2-4} 

 & Acoustic \newline Event \newline Recognition 
 & Identify and classify the type of a sound in the video, based on both visual and audio cues. 
 & From the moment a wolf picks up a green pot on the ground and shows it to the little girl until the two wolves are blown away, which of the following sounds was not heard?  \newline Options: \newline A. Pig squealing \newline \textcolor{green!50!black}{B. Wolf howling} \newline C. Fly buzzing \newline D. Dog barking  \\
\hline

  \textbf{Perception}& Acoustic \newline Attribute \newline Recognition 
 & Identify the acoustic attributes of specific acoustic events, based on both visual and audio cues. 
 & Compared to the background music used when the vlogger was cleaning, what are the characteristics of the background music during the sushi dinner scene?  \newline Options: \newline A. Slower tempo \newline B. Louder volume \newline \textcolor{green!50!black}{C. Faster tempo} \newline D. Softer volume  \\
\cline{2-4} 

& Speech \newline Fact \newline Retrieval 
 & Retrieve explicitly mentioned factual information from speech to answer questions, based on both visual and audio cues. 
 & In the beginning of class, what feature of the map provided by the teacher did Abby point out in her first comment?  \newline Options: \newline A. There is hilly terrain. \newline \textcolor{green!50!black}{B. There is a graveyard.} \newline C. There is a train. \newline D. There is a factory.  \\
\hline

\textbf{Perception} & Captioning 
 & Generate a descriptive text that details the observable visual actions and audible acoustic events, including speech, environmental sounds, and scene dynamics, occurring in the video. 
 & Describe the content in the video between 01:02:38 and 01:02:47, including all audio and visual information.  \newline Options: \newline \textcolor{green!50!black}{A. Amidst the background music, on a night of heavy rain with the sound of pouring rain audible, a bear leans out of the window and looks up at the creaking, wobbling antenna on the roof. It climbs onto the roof and grabs the antenna. Then, a flash of lightning appears in the sky, accompanied by the sound of thunder—a bolt of lightning strikes the antenna, electrocuting the bear, which lets out a cry along with the sound of the electric shock.} \newline B. On a stormy night with background music and the sound of heavy rain, a bear climbs from a window onto the roof, looking at the creaking antenna. A sudden flash of lightning illuminates the sky, followed by a loud clap of thunder. Startled by the noise, the bear loses its footing and grabs the wobbling antenna to steady itself. At that moment, it is electrocuted, letting out a pained cry accompanied by the sound of an electric shock. \newline C. Amidst the background music on a night of pouring rain, a bear is seen climbing onto the roof to deal with a creaking, wobbling antenna. As it grabs the metal pole, a strong gust of wind blows it over, causing it to make contact with a live power line. A surge of electricity courses through the bear, which is instantly electrocuted and lets out a loud cry as the sound of a powerful electric shock is heard over the rain and thunder. \newline D. On a night filled with background music and the sound of a thunderstorm, a bear climbs onto the roof and grabs the creaking, wobbling antenna. As it holds on, a massive bolt of lightning strikes the chimney right next to it, accompanied by a deafening clap of thunder. Terrified by the close strike, the bear lets go of the antenna, loses its footing on the wet roof, and silently falls off, with only the storm audible.  \\
\hline

\textbf{Cognition}& Causal \newline Reasoning  
 & Infer the underlying causes or resulting consequences of a specific event, based on both visual and audio cues. 
 & What sound caused the two wolves with pacifiers to wake up in the crib at night?  \newline Options: \newline \textcolor{green!50!black}{A. Elephant call} \newline B. Pig oinking \newline C. Explosion sound \newline D. Scream  \\
\cline{2-4} 

& Emotional \newline Reasoning 
 & Infer the emotional state, underlying causes, and evolutionary trajectories of specific entities or the overall atmosphere, based on both visual and audio cues. 
 & When Christian's parents had their second conversation, what kind of atmosphere did the background music convey?  \newline Options: \newline A. Tense \newline B. Relaxed \newline C. Cheerful \newline \textcolor{green!50!black}{D. Romantic}  \\
\cline{2-4} 

\textbf{Cognition}& Intentional \newline Reasoning 
 & Infer the underlying purposes or motivations behind a specific character’s actions, based on both visual and audio cues. 
 & When a bear was about to go out, a little girl rode her bike and opened the door to enter its home. What was the little girl's purpose for coming here?  \newline Options: \newline \textcolor{green!50!black}{A. Have the bear provide breakfast for her.} \newline B. Show the bear her performance. \newline C. Eat with the bear. \newline D. Ask the bear for cooking ingredients.  \\
\cline{2-4} 

& Object \newline Reasoning 
 & Infer a specific object that meets a certain condition, or infer its function, attributes, or relationships between objects, based on both visual and audio cues. 
 & Based on students' comments and behaviors during the first 45 minutes of class, which of the following students do you think was the most active in class?  \newline Options: \newline \textcolor{green!50!black}{A. Abby} \newline B. Dylan \newline C. Eric \newline D. Bernadette  \\
\cline{2-4} 

& Speech \newline Content \newline Reasoning 
 & Reason about implicit content that is not explicitly mentioned in the speech to answer questions, based on both visual and audio cues. 
 & Why does a man in a brown shirt and a man in a blue top appear somewhat angry at a man in an orange top and turn away together?  \newline Options: \newline \textcolor{green!50!black}{A. Because the man in the orange top told the woman in the purple dress about his intention to propose before informing them.} \newline B. Because the man in the orange top told the woman in the purple dress about his wedding plans before informing them. \newline C. Because the man in the orange top told the woman in the green top about his intention to propose before informing them. \newline D. Because the man in the orange top told the woman in the green top about his wedding plans before informing them.  \\
\cline{2-4} 

& Temporal \newline Ordering 
 & Arrange multiple key events from the video according to both visual and audio cues. 
 & Please arrange the following students in chronological order based on the time of their last appearance in the video: 1. the girl in the purple top, 2. the boy wearing khaki shorts, 3. the boy in the white hoodie, 4. the girl in the pink top.  \newline Options: \newline \textcolor{green!50!black}{A. 1432} \newline B. 1342 \newline C. 1234 \newline D. 1243  \\
\cline{2-4} 

& Spatial \newline Reasoning 
 & Analyze the audio-visual information to reason about the spatial locations of objects. 
     & When the main photographer first began taking photos of the bride and groom during their vows, where was Ann located relative to the main photographer?  \newline Options: \newline A. Directly in left \newline B. Back \newline \textcolor{green!50!black}{C. Right-front} \newline D. Left-front  \\
\hline

\textbf{Summarization}& Summarization 
 & Analyze the audio-visual information to achieve a high-level, abstract understanding. 
     & What was the topic of the question asked by the female student wearing black clothing, with brown hair, and facing away from the camera, to the guest speaker on stage?  \newline Options: \newline A. The challenge of lacking computational resources for training \newline \textcolor{green!50!black}{B. The impact of artificial intelligence on education} \newline C. Criteria for admitting students to their lab \newline D. Collaboration between private companies and universities  \\
\hline

\textbf{Temporal \newline Grounding}& Temporal \newline Grounding 
 & Locate the timestamp of specific audio-visual events. 
     & Please locate the timestamps of Ghufaira's first and last appearances in the video.  \newline Options: \newline \textcolor{green!50!black}{A. 00:02:16-01:44:56} \newline B. 00:01:16-01:45:56 \newline C. 00:12:16-01:34:56 \newline D. 00:13:16-01:35:56 \\
\hline

\end{longtable}

\newpage
\section{More statistics of our dataset}
\label{appendix_c}

\begin{figure}[h] 
    \centering 
    \includegraphics[width=0.7\textwidth]{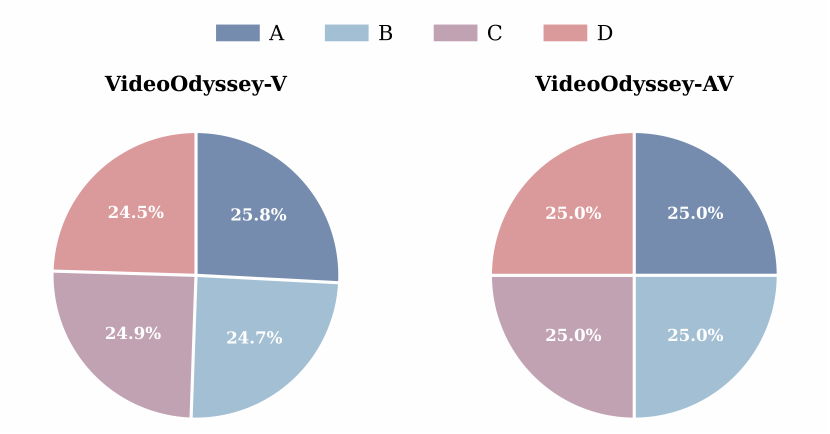} 
    \caption{Answer distribution of \textit{VideoOdyssey-V} and \textit{VideoOdyssey-AV}} 
    \label{fig:answer distibution} 
\end{figure}

Fig. \ref{fig:answer distibution} illustrates the balanced answer distribution for both \textit{VideoOdyssey-V} and \textit{VideoOdyssey-AV}. Table \ref{tab:overall_statistics} presents a comprehensive statistic overview of the \textit{VideoOdyssey} benchmark. Furthermore, Table \ref{tab:v_num_avg_CCL_across_task_types} and Table \ref{tab:av_num_avg_CCL_across_task_types} detail the number of QA pairs and the average continuous certificate length (CCL) across different task types in the \textit{VideoOdyssey-V} and \textit{VideoOdyssey-AV}, respectively. Finally, Table \ref{tab:av_num_avg_CCL_across_audio_types} show the number of QA pairs and the average continuous certificate length corresponding to the three audio types within \textit{VideoOdyssey-AV}.

\begin{table}[h]
\caption{Statistics of our dataset}
\begin{adjustbox}{max width=1.0\textwidth}
  \centering
  \begin{tabular}{ll ll ll}
    \toprule
    \multicolumn{2}{c}{\textbf{Video Statistics}} & \multicolumn{2}{c}{\textbf{\textit{VideoOdyssey-V}}} & \multicolumn{2}{c}{\textbf{\textit{VideoOdyssey-AV}}} \\
    \midrule
    \#Videos & 100 & \#Task Types & 14 & \#Task Types & 18\\
    \#Major Categories & 11  & \#Questions & 1618 & \#Questions & 1062  \\
    \#Subcategories    & 54       & Avg. Question Len. & 21.73 words  & Avg. Question Len. & 21.67 words \\
    Duration      & 60-255min & Avg. Option Len.& 5.54 words  & Avg. Option Len.& 7.11 words \\
    Avg. Duration      & 109min     & Avg. CCL&  16.0 min & Avg. CCL&  12.8 min\\
    Audio Types & Sp+So+Mu & - & - & Audio Types (Sp:So:Mu) & 743:258:149 \\ 
    -&-& Multi-level CCL & 539:382:288:272:137 & Multi-level CCL &244:341:272:133:72\\
    -&-& Avg. Question per Hour & 8.91 & Avg. Question per Hour  & 5.85\\

    \bottomrule
  \end{tabular}
  \end{adjustbox}
  \label{tab:overall_statistics}
\end{table}

\begin{table}[H]
  \centering
  \caption{Numbers and average continuous certificate length (CCL) across different task types in \textit{VideoOdyssey-V}.}
  \vspace{+0.1cm}
  \begin{adjustbox}{max width=1.0\textwidth}
    \setlength{\tabcolsep}{3pt} 
    \begin{tabular}{l *{14}{c}} 
      \toprule
      & Order & Count & ObRea & TeGro & OCR & SpRea & ObRec & Sum & AtRec & EmRea & InRea & AcRec & CaRea &Cap \\
      \midrule
      Numbers                      & 89 & 307 & 173& 65 & 167 & 61 & 208 & 65 & 106 & 71 & 66 & 192 & 128 & 49   \\
      Avg. CCL (min) & 32.3 & 31.4 & 26.3 & 17.1 & 13.8 & 12.8 & 12.2 & 11.9 & 9.2 & 7.1 & 5.9 & 5.4 & 4.4 & 1.3  \\
      \bottomrule
    \end{tabular}
  \end{adjustbox}
  \label{tab:v_num_avg_CCL_across_task_types}
\end{table}

\begin{table}[H]
  \centering
  \caption{Numbers and average continuous certificate length (CCL) across different task types in \textit{VideoOdyssey-AV}.}
  \vspace{+0.1cm}
  \begin{adjustbox}{max width=1.0\textwidth}
    \setlength{\tabcolsep}{2.5pt} 
    \begin{tabular}{l *{18}{c}} 
      \toprule
      & Order & TeGro & Count & OCR & ObRea & ObRec & VAR & AER & Cap & AAR & EmRea & Sum & SpRea & AcRec & InRea & CaRea & SCR & SFR \\
      \midrule
      Numbers    & 50&61&81&48&56&63&58&86&47&65&62&50&60&60&52&58&62&173   \\
      Avg. CCL (min) & 34.6&21.6&21.1&19.1&18.8&17.9&14.2&13.5&12.8&11.4&11.3&11.1&10.1&9.5&9.1&8.8&8.3&7.4 \\
      \bottomrule
    \end{tabular}
  \end{adjustbox}
  \label{tab:av_num_avg_CCL_across_task_types}
\end{table}

\begin{table}[H]
  \centering
  \caption{Numbers and average continuous certificate length (CCL) across different audio types in \textit{VideoOdyssey-AV}.}
  \vspace{+0.1cm}
  \begin{adjustbox}{max width=0.42\textwidth}
    \begin{tabular}{l *{3}{c}} 
      \toprule
      & Speech &Sound&Music\\
      \midrule
      Numbers    & 743 &258&149  \\
      Avg. CCL (min) & 13.1&11.2&10.7 \\
      \bottomrule
    \end{tabular}
  \end{adjustbox}
  \label{tab:av_num_avg_CCL_across_audio_types}
\end{table}


\subsection{Source-level robustness and within-video diversity}
\label{app:source_robustness}
Within-video QA pairs are not entirely independent, but no individual video disproportionately determines the aggregate result.

We further removed each of the 100 videos in turn and recomputed accuracy. The small maximum changes in Table~\ref{tab:leave_one_video_out} show that no single source video dominates the reported overall result.

\begin{table}[htbp]
\centering
\small
\caption{Leave-one-video-out accuracy (\%) and maximum change (percentage points).}
\label{tab:leave_one_video_out}
\begin{adjustbox}{max width=\textwidth}
\begin{tabular}{llccc}
\toprule
Model & Subset & Original accuracy & \shortstack{Leave-one-video-out\\accuracy range} & \shortstack{Maximum change\\(pp)} \\
\midrule
Gemini-3.1-Pro & V & 56.3 & 55.8--57.0 & 0.7 \\
Qwen3.5-27B & V & 44.6 & 44.1--44.9 & 0.5 \\
Gemini-3.1-Pro & AV & 46.1 & 45.5--46.5 & 0.6 \\
Qwen3-Omni-30B & AV & 28.7 & 28.3--29.1 & 0.4 \\
\bottomrule
\end{tabular}
\end{adjustbox}
\end{table}

\paragraph{Within-video diversity.}
We also audited within-video diversity. The median number of task types is 6; 90\% of V and 85\% of AV videos contain at least 3 types. Dividing each video into 4 equal segments by CCL-window midpoint, 84\% and 66\% cover at least 3 segments, while 78.8\% and 74.3\% of QA-window pairs do not overlap. Thus, same-video questions generally span diverse tasks and times rather than repeatedly querying one event.

\paragraph{Collection scope and potential contamination.}
Dynamism was screened manually rather than by a quantitative metric: annotators excluded videos composed primarily of static images, slides, or long periods without meaningful content changes.

Because the videos are public, we cannot completely exclude pretraining exposure. To reduce knowledge-based shortcuts, recognizable TV/Movie questions avoid character/actor names and well-known relationships or plots recoverable from public summaries. Characters are instead described by visible clothing and appearance, and questions require video-specific evidence. Together with language-only shortcut filtering and human review, this reduces textual and world-knowledge shortcuts.

\clearpage

\section{More results}
\label{appendix_d}

\subsection{Performance across video domains}
\begin{figure}[t] 
    \centering 
    \includegraphics[width=1.0\textwidth]{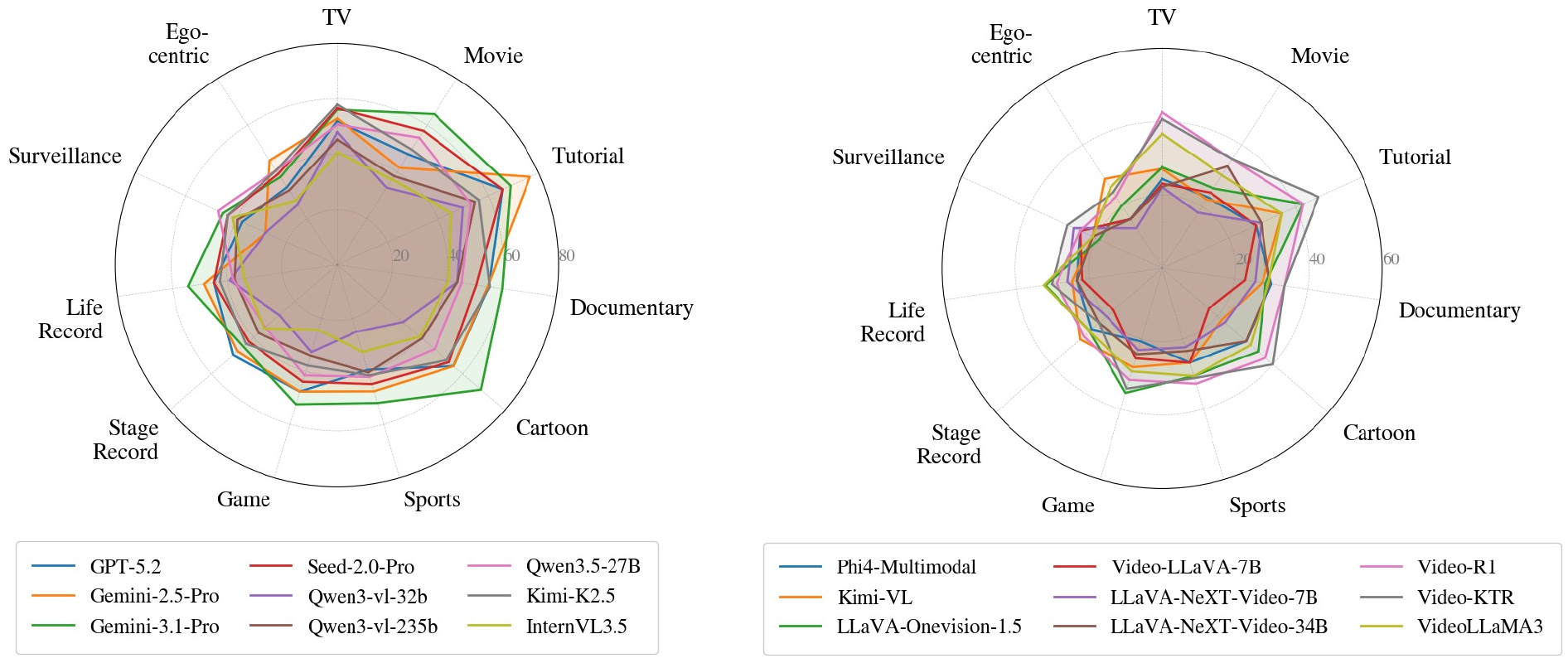} 
    \caption{Performancewa across different video domains on \textit{VideoOdyssey-V}.} 
    \label{fig:v_performance_across_video_domains} 
\end{figure}

\begin{figure}[t] 
    \centering 
    \includegraphics[width=1.0\textwidth]{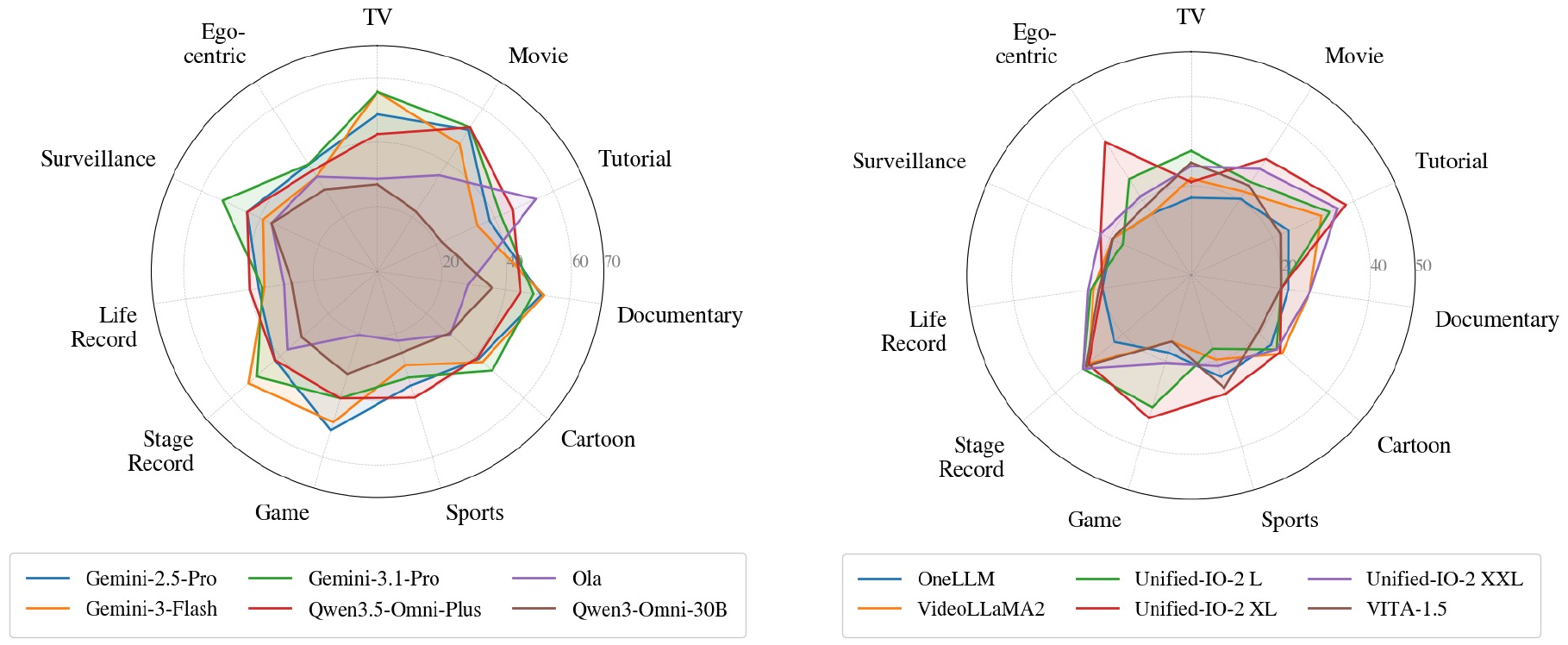} 
    \caption{Performancewa across different video domains on \textit{VideoOdyssey-AV}.} 
    \label{fig:av_performance_across_video_domains} 
\end{figure}
The radar charts (Fig. \ref{fig:v_performance_across_video_domains} and Fig. \ref{fig:av_performance_across_video_domains}) expose a pronounced domain bias. In pure visual tasks, models consistently excel on highly structured content (\eg, TV, Tutorial, Movie) while struggling with unstructured domains (\eg, Surveillance, Ego-centric videos). 
Transitioning to the audio-visual setting introduces significant performance volatility. Despite this instability, models maintain a pronounced advantage in domains characterized by rich speech cues and structured content (\eg, TV, Movie, Documentary), whereas Ego-centric scenarios remain overwhelmingly challenging.

\subsection{Impact of input modalities}

\begin{figure}[t] 
    \centering 
    \includegraphics[width=1.0\textwidth]{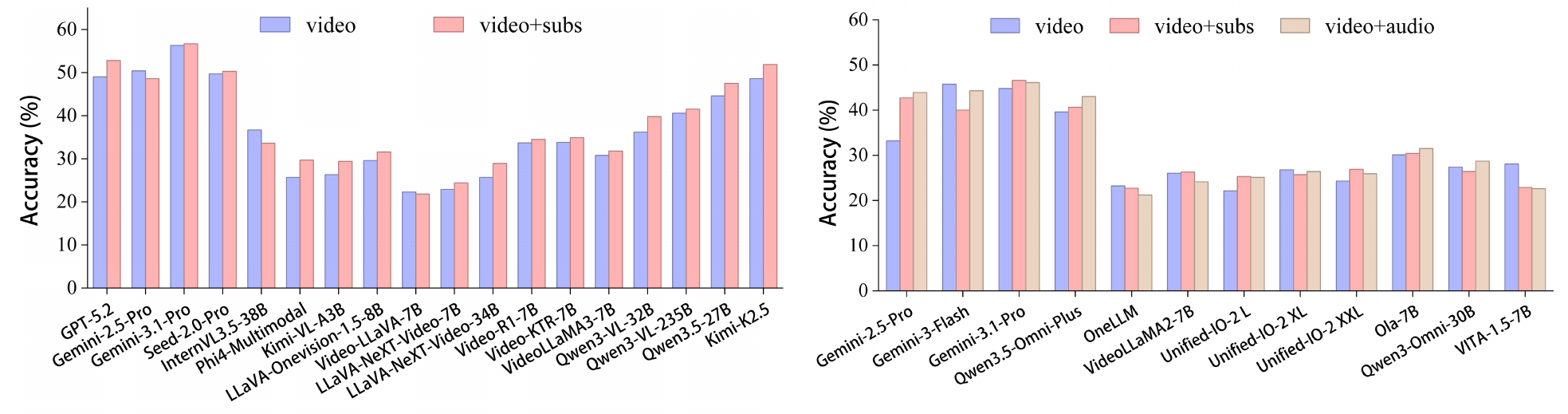} 
    \caption{Performance of MLLMs under different input modalities on \textit{VideoOdyssey-V} \textbf{(left)} and \textit{VideoOdyssey-AV} \textbf{(right)}}
    \label{fig:impact_of_modality_overall}
\end{figure}

Fig. ~\ref{fig:impact_of_modality_overall} illustrates the impact of different input modalities on overall model accuracy. In pure visual tasks, adding subtitles generally boosts performance by providing complementary semantic context. However, exceptions like Gemini-2.5-Pro, InternVL3.5-38B, and Video-LLaVA-7B suffer performance drops, likely due to a "text-bias" where over-reliance on linguistic input distracts from visual reasoning.

In audio-visual tasks, however, richer modalities do not guarantee better accuracy. For several models, extra modalities act as distractors and degrade performance, highlighting deficiencies in cross-modal alignment. Strikingly, despite its strong overall performance in main evaluations, Gemini-3-Flash peaks with purely visual input and degrades once audio or subtitles are introduced, revealing that its high accuracy does not stem from effective cross-modal comprehension. Meanwhile, other models (e.g., Gemini-3.1-Pro and Unified-IO-2 XXL) peak with video+subtitles, exposing a heavy reliance on text rather than true acoustic comprehension. Conversely, robust native omni-modal architectures (i.e., Gemini-2.5-Pro, Qwen3.5-Omni-Plus, Ola-7B, and Qwen3-Omni-30B) successfully integrate and benefit from the audio stream, achieving peak accuracy under the video+audio setting.

\subsection{Human performance across continuous certificate length levels}

\begin{figure}[t] 
    \centering 
    \includegraphics[width=0.9\textwidth]{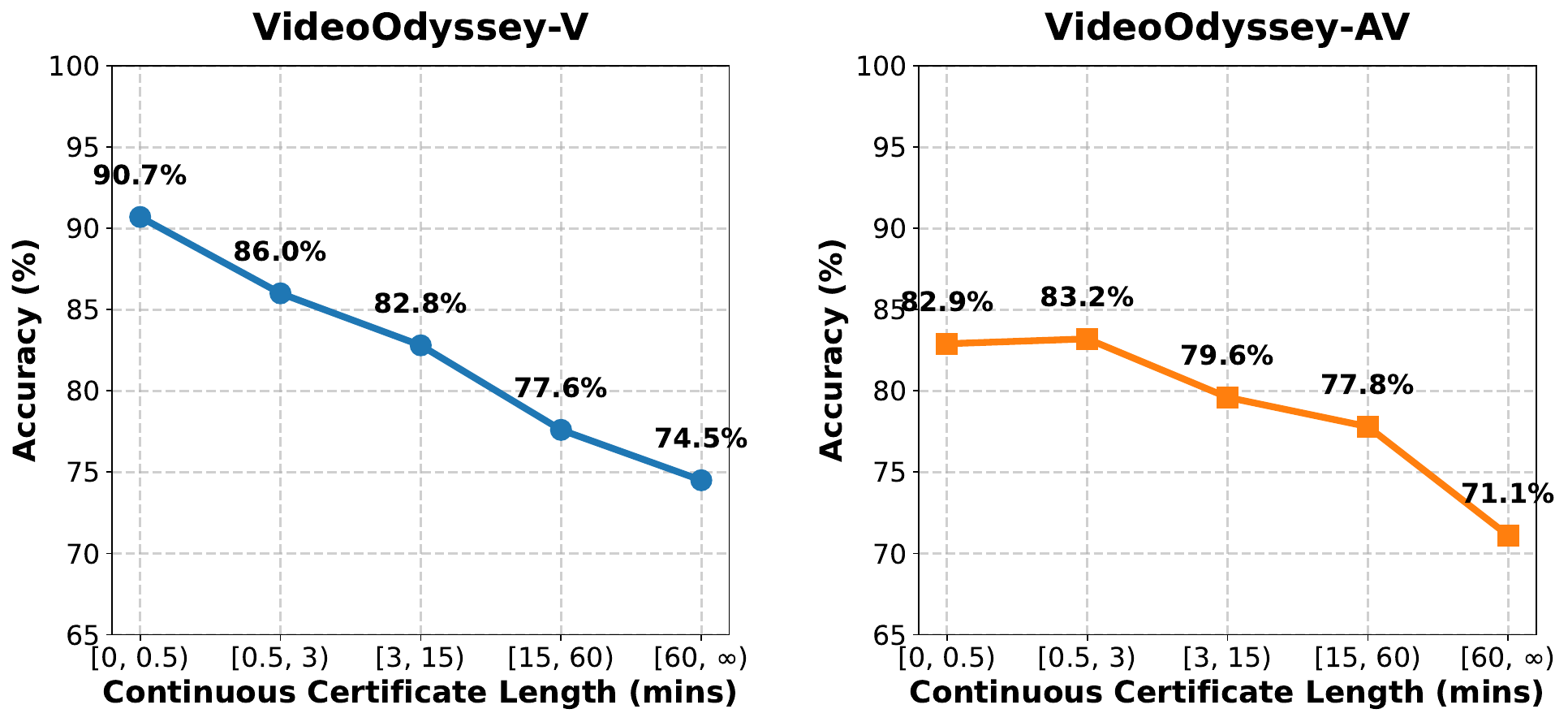} 
    \caption{Human performance across continuous certificate length levels}
    \label{fig:human_performance_ccl}
\end{figure}

To further understand the correlation between the temporal span of evidence and human cognitive load, we present the human performance across varying Continuous Certificate Length (CCL) levels on both \textit{VideoOdyssey-V} and \textit{VideoOdyssey-AV} settings. (Fig. \ref{fig:human_performance_ccl})

As illustrated in the figures, there is a clear negative correlation between the continuous certificate length and human accuracy. In \textit{VideoOdyssey-V}, human performance starts at a high of 90.7\% for extremely short evidence lengths ($<0.5$ mins) but drops strictly and significantly to 74.5\% for evidence lengths exceeding 60 minutes. A similar declining trend is observed in \textit{VideoOdyssey-AV}, dropping from around 83\% to 71.1\%. Performance across CCL bins therefore diagnoses behavior by required continuous temporal span; it does not imply that longer CCL alone lowers accuracy.

\subsection{Human unimodal ablations on VideoOdyssey-AV}
\label{app:human_unimodal}
We conducted forced-choice human unimodal ablations on VideoOdyssey-AV. Human accuracy is 32.4\% with video only and 33.3\% with audio only, both close to the 25\% chance level. These results show that neither modality alone provides sufficient evidence for reliable answering, supporting the need for audio-visual integration.

\subsection{Sensitivity to frame budgets}
\label{app:frame_budget_sensitivity}
We additionally test two frame budgets (Table~\ref{tab:frame_budget_sensitivity}). Increasing frames has model-dependent effects rather than a consistent gain: sampling remains sparse for multi-hour videos, while additional frames may add irrelevant content and complicate evidence selection.

The effects are model-dependent: Gemini-2.5-Pro changes from 51.1\% to 50.4\%, Qwen3-VL-235B from 38.1\% to 40.6\%, Gemini-3-Flash from 45.6\% to 44.3\%, and Qwen3-Omni-30B from 31.2\% to 28.7\%.

\begin{table}[htbp]
\centering
\small
\caption{Frame-budget sensitivity under full-video input (w/o CW). All scores are accuracy (\%); CCL intervals are in minutes.}
\label{tab:frame_budget_sensitivity}
\begin{adjustbox}{max width=\textwidth}
\setlength{\tabcolsep}{4pt}
\begin{tabular}{lccccccc}
\toprule
Model & Frames & Overall & $[0,0.5)$ & $[0.5,3)$ & $[3,15)$ & $[15,60)$ & $[60,\infty)$ \\
\midrule
\multicolumn{8}{c}{\textbf{(a) VideoOdyssey-V}} \\
\midrule
Gemini-2.5-Pro & 64 & 51.1 & 49.2 & 52.1 & 53.8 & 51.1 & 50.7 \\
Gemini-2.5-Pro & 128 & 50.4 & 47.1 & 49.5 & 53.8 & 54.4 & 50.4 \\
Qwen3-VL-235B & 64 & 38.1 & 37.3 & 40.6 & 39.6 & 38.6 & 30.7 \\
Qwen3-VL-235B & 128 & 40.6 & 41.0 & 42.7 & 43.1 & 39.0 & 31.4 \\
\midrule
\multicolumn{8}{c}{\textbf{(b) VideoOdyssey-AV}} \\
\midrule
Gemini-3-Flash & 32 & 45.6 & 47.1 & 45.8 & 47.2 & 44.5 & 43.6 \\
Gemini-3-Flash & 64 & 44.3 & 44.7 & 41.4 & 47.1 & 44.4 & 45.8 \\
Qwen3-Omni-30B & 32 & 31.2 & 31.6 & 29.0 & 36.7 & 28.6 & 27.8 \\
Qwen3-Omni-30B & 64 & 28.7 & 27.1 & 27.9 & 28.3 & 29.3 & 38.9 \\
\bottomrule
\end{tabular}
\end{adjustbox}
\end{table}

\paragraph{Sampling protocol and coverage.}
We use deterministic uniform sampling: for $F$ video frames and budget $N$, we sample $N$ frames at a fixed interval of $F/N$. The same video and budget always produce the same input.

Our deterministic uniform sparse sampling follows common long-video MLLM evaluation practice, but can miss rapid, fine-grained events. For videos averaging 109 minutes, budgets of 64/128/256 sample approximately one frame every 102/51/26 seconds; complete coverage of every transient detail is therefore impossible.

\section{Details of measuring continuous certificate length}
\label{appendix_e}

We manually measured the average continuous certificate length for the 17 datasets featured in Table \ref{tab:dataset_comparison}, including MovieChat-1K~\citep{song2024moviechat}, LongVideoBench~\citep{wu2024longvideobench}, Video-MMMU~\citep{hu2025video}, MLVU~\citep{zhou2025mlvu}, Video-MME~\citep{fu2025video}, CG-Bench~\citep{chen2024cg}, InfiniBench~\citep{ataallah2025infinibench}, LVBench~\cite{wang2025lvbench}, AVQA~\citep{yang2022avqa}, Music-AVQA~\citep{li2022learning}, LongVALE~\citep{geng2025longvale}, Daily-Omni~\citep{zhou2025daily}, OmniVideoBench~\citep{li2025omnivideobench}, LongInsightBench~\citep{han2025longinsightbench}, WorldSense~\citep{hong2025worldsense}, LVOmniBench~\citep{tao2026lvomnibench}, Video-MME-v2~\citep{fu2026video}. Given the labor-intensive nature of identifying temporal evidence, we dedicated over 4 hours of manual annotation to each individual dataset.

We randomly sampled 100 QA pairs from each benchmark and carefully ensured that the selected subsets comprehensively covered all predefined question types, as well as all video duration groups for datasets that categorize videos by duration (\ie, LongVideoBench and Video-MME).
Additionally, during the annotation process for the WorldSense dataset, we observed that the sampled subset contains both audio-visual queries and single-modality (i.e., visual-only or audio-only) queries. Therefore, we directly categorized it as a "mixed" dataset in Table \ref{tab:dataset_comparison} to accurately reflect its practical composition.

Based on our measurements, \textit{VideoOdyssey} extend the average continuous certificate to an unprecedented 16 minutes for \textit{VideoOdyssey-V} and 12.8 minutes for \textit{VideoOdyssey-AV}. Compared to the existing benchmarks with the longest video durations, we increase this metric by 4 times and 16 times for the pure visual (LVBench) and audio-visual (LVOmniBench) domains, respectively.

\subsection{CCL annotation protocol and independent re-annotation}
\label{app:ccl_reliability}
The initial QA pairs and CCL windows were annotated by 14 uniformly trained annotators. For each question, the annotator identified all necessary evidence and marked the shortest continuous interval containing it. A local cue retained only its shortest clearly recognizable segment; distributed evidence extended from the earliest necessary evidence to the end of the latest. Both boundaries were tightened until further trimming would remove necessary evidence or prevent a unique answer.

Each question was then checked by a reviewer independent of its annotator. Using only the annotated window, the reviewer verified that the answer was correct, uniquely determined, and supported within the window, and checked for redundant or shifted CCL boundaries. Noncompliant windows or answers were revised; samples that could not meet the objectivity and clarity requirements were removed.

On V subset, 12 windows were adjusted and 6 crossed a CCL-bin boundary. On AV subset, 9 windows were adjusted and 5 crossed a boundary.

The new independent re-annotation results show high CCL stability (Table~\ref{tab:ccl_reannotation}). Among the 35 V bin-level disagreements, 30 occur at bin boundaries. All 22 AV bin-level disagreements occur at bin boundaries.

\begin{table}[htbp]
\centering
\small
\caption{Independent CCL re-annotation results on all 1,618 V samples and all 1,062 AV samples.}
\label{tab:ccl_reannotation}
\begin{adjustbox}{max width=\textwidth}
\begin{tabular}{lrrrrr}
\toprule
Subset & \shortstack{Total\\questions} & \shortstack{Same-bin\\questions} & \shortstack{Bin agreement\\(\%)} & \shortstack{Mean\\window IoU} & \shortstack{Median\\window IoU} \\
\midrule
VideoOdyssey-V & 1,618 & 1,583 & 97.8 & 0.932 & 1.0 \\
VideoOdyssey-AV & 1,062 & 1,040 & 97.9 & 0.922 & 1.0 \\
\bottomrule
\end{tabular}
\end{adjustbox}
\end{table}

\paragraph{Interpretation of CCL.}
CCL is a diagnostic dimension, not a complete or causal measure of difficulty. It is defined independently of model performance as the continuous temporal span containing the required evidence. Difficulty also depends on reasoning type, distractors, modality and audio type, so equal-CCL questions may differ substantially. Performance across CCL bins therefore diagnoses behavior by required continuous temporal span; it does not imply that longer CCL alone lowers accuracy.

\subsection{Annotation corrections across temporal scales}
\label{app:annotation_corrections}
The recorded corrections do not increase monotonically with either CCL or video duration. We analyze the complete original annotations and revision records; the rates in Table~\ref{tab:annotation_corrections} count answer corrections among retained QA pairs, while deleted questions are reported separately.

\begin{table}[htbp]
\centering
\small
\caption{Answer-correction rates (\%) among retained QA pairs, grouped by (a) CCL and (b) full-video duration. All temporal intervals are in minutes.}
\label{tab:annotation_corrections}
\textbf{(a) Grouped by CCL}\par\smallskip
\begin{tabular}{lccccc}
\toprule
Subset & $[0,0.5)$ & $[0.5,3)$ & $[3,15)$ & $[15,60)$ & $[60,\infty)$ \\
\midrule
VideoOdyssey-V & 2.1 & 2.6 & 1.0 & 1.1 & 2.1 \\
VideoOdyssey-AV & 4.9 & 2.9 & 3.3 & 4.5 & 0.0 \\
\bottomrule
\end{tabular}
\par\medskip
\textbf{(b) Grouped by full-video duration}\par\smallskip
\begin{tabular}{lcccc}
\toprule
Subset & $[60,90)$ & $[90,120)$ & $[120,180)$ & $[180,\infty)$ \\
\midrule
VideoOdyssey-V & 3.6 & 2.4 & 2.8 & 1.1 \\
VideoOdyssey-AV & 3.0 & 4.7 & 3.2 & 3.4 \\
\bottomrule
\end{tabular}
\end{table}

During quality control, 46 V and 79 AV QA pairs were removed. Among retained pairs, 27 V answers were corrected (1.7\%; 12 Counting and 6 Object Recognition), while 37 AV answers were corrected (3.5\%; 9 Acoustic Event Recognition, 7 Counting, 7 Object Recognition, and 5 Visual Attribute Recognition). So we do not infer that these tasks are intrinsically more error-prone.

\section{Details for evaluation with ground-truth certificate window}
\label{appendix_f}

The original w/ CW setting combines evidence localization with denser local sampling and was not intended as a strictly controlled comparison. It is an idealized setting that directly provides the ground-truth window and samples the maximum supported frames, measuring performance when localization and input budget are favorable. Specifically, we adopt a dynamic sampling strategy. Taking Gemini-2.5-Pro as an example, the model is configured to process a maximum of 128 frames in Table \ref{tab:visual_main}. When evaluating this model using the ground-truth certificate window, the sampling strategy is as follows:
\begin{itemize}
    \item Window length < 128 seconds: We densely extract frames at a rate of 1fps to preserve maximum temporal granularity.
    \item Window length $\geq$ 128 seconds: We uniformly sample 128 frames across the entire duration of the window to provide a comprehensive overview of the localized event.
\end{itemize}

This dynamic sampling strategy is consistently applied across all evaluated models. 

\subsection{Matched-rate certificate-window evaluation}
\label{app:cw_matched_rate}
To disentangle localization from density, we conduct a matched-sampling-rate experiment. It uses the full-video temporal sampling interval: for video duration $T$, window duration $L$, and full-video frame budget $N$, it samples
\[
N_{\mathrm{CW}} = \max\left(1,\mathrm{round}\left(\frac{NL}{T}\right)\right)
\]
frames. Table~\ref{tab:cw_sampling_controls} reports the results.

\begin{table}[H]
\centering
\small
\caption{Full-video, matched-frame CW, and matched-rate CW accuracy (\%). CCL intervals are in minutes; $N$ denotes the full-video frame budget.}
\label{tab:cw_sampling_controls}
\begin{adjustbox}{max width=\textwidth}
\setlength{\tabcolsep}{5pt}
\begin{tabular}{lcccccc}
\toprule
Setting & Overall & $[0,0.5)$ & $[0.5,3)$ & $[3,15)$ & $[15,60)$ & $[60,\infty)$ \\
\midrule
\multicolumn{7}{l}{\textbf{Gemini-2.5-Pro (V), $N=128$}} \\
Full video (w/o CW) & 50.4 & 47.1 & 49.5 & 53.8 & 54.4 & 50.4 \\
CW, matched frames & 59.8 & 67.9 & 72.3 & 56.9 & 38.6 & 40.9 \\
CW, matched rate & 53.7 & 50.1 & 55.2 & 59.1 & 54.0 & 48.2 \\
\midrule
\multicolumn{7}{l}{\textbf{Qwen3-VL-235B (V), $N=128$}} \\
Full video (w/o CW) & 40.6 & 41.0 & 42.7 & 43.1 & 39.0 & 31.4 \\
CW, matched frames & 56.9 & 61.0 & 67.0 & 60.1 & 48.5 & 31.7 \\
CW, matched rate & 42.5 & 41.9 & 44.0 & 47.2 & 38.7 & 38.6 \\
\midrule
\multicolumn{7}{l}{\textbf{Gemini-3-Flash (AV), $N=64$}} \\
Full video (w/o CW) & 44.3 & 44.7 & 41.4 & 47.1 & 44.4 & 45.8 \\
CW, matched frames & 55.5 & 68.9 & 57.8 & 52.6 & 44.4 & 30.6 \\
CW, matched rate & 53.4 & 67.6 & 54.0 & 48.2 & 46.6 & 34.7 \\
\midrule
\multicolumn{7}{l}{\textbf{Qwen3-Omni-30B (AV), $N=64$}} \\
Full video (w/o CW) & 28.7 & 27.1 & 27.9 & 28.3 & 29.3 & 38.9 \\
CW, matched frames & 48.6 & 61.1 & 53.7 & 44.5 & 29.3 & 33.3 \\
CW, matched rate & 38.6 & 46.7 & 41.4 & 34.6 & 30.1 & 29.2 \\
\bottomrule
\end{tabular}
\end{adjustbox}
\end{table}

After controlling for denser local sampling, providing the ground-truth evidence window improves overall accuracy for every tested model: Gemini-2.5-Pro from 50.4\% to 53.7\%, Qwen3-VL-235B from 40.6\% to 42.5\%, Gemini-3-Flash from 44.3\% to 53.4\%, and Qwen3-Omni-30B from 28.7\% to 38.6\%. This shows that the CW gain is not entirely caused by higher sampling density. Fixed-frame CW obtains 59.8\%, 56.9\%, 55.5\%, and 48.6\%, respectively, confirming that the original results also contain gains from denser local sampling.

For three of the four models, matched-rate gains concentrate in short-to-medium CCL bins and accuracy decreases in the longest bin; Qwen3-VL-235B is the exception. We thus retain only the directly supported conclusion that matched-rate CW does not yield a consistent improvement in the longest CCL range.

\paragraph{Scope of the input protocol.}
Some models support native video, but their APIs still apply provider-side preprocessing and media-resolution/context limits; with text and audio, some ultra-long videos cannot fit. Native video is also unavailable for most evaluated models and costly at this scale, so it would cover a non-comparable subset of models and videos. We therefore use reproducible uniform frame sampling for the main comparison.

\section{Evaluation prompts}
\label{appendix_g}

When the input consists solely of the video, we follow the following prompts:

\begin{figure}[h] 
    \centering 
    \includegraphics[width=1.0\textwidth]{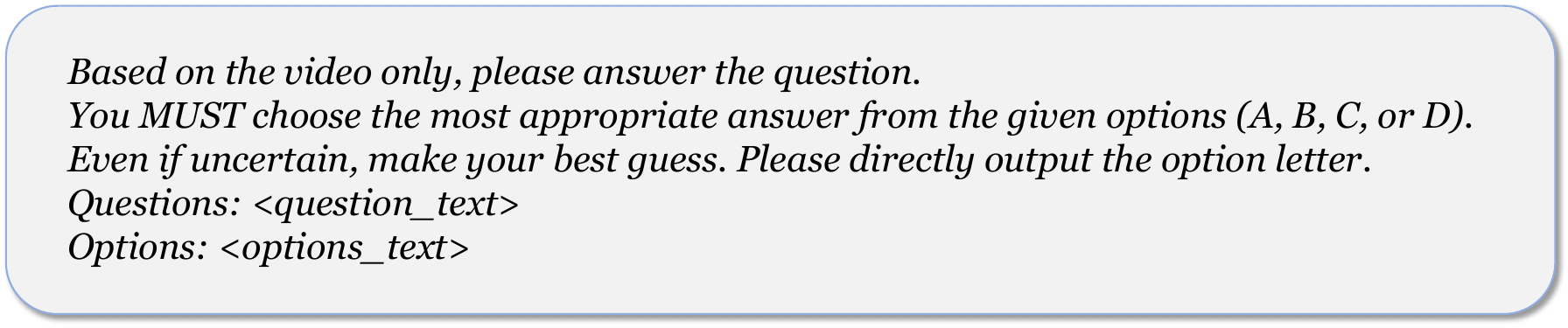} 
    
    \label{fig:prompt_v}
\end{figure}

When both the video and subtitles are provided as input, we follow the following prompts:

\begin{figure}[H] 
\vspace{-0.2cm}
    \centering 
    \includegraphics[width=1.0\textwidth]{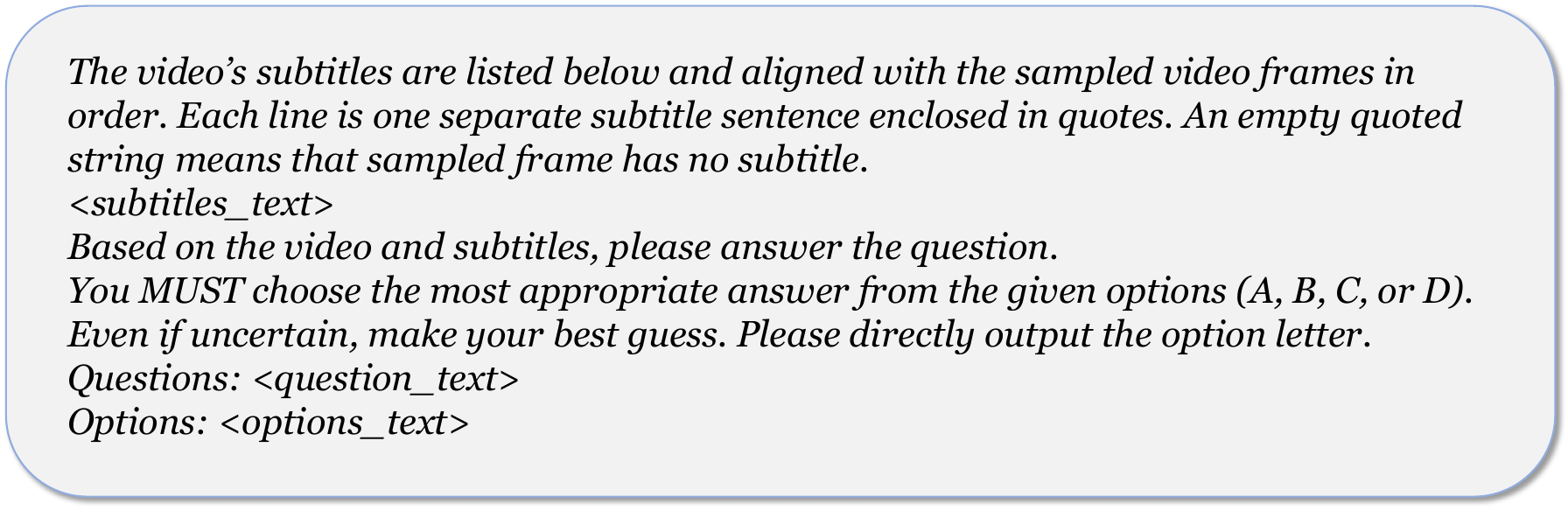} 
    \vspace{-0.5cm}
   
    \label{fig:prompt_vs}
\end{figure}

When the input includes both the video and the audio track, we follow the following prompts:

\begin{figure}[H] 
\vspace{-0.2cm}
    \centering 
    \includegraphics[width=1.0\textwidth]{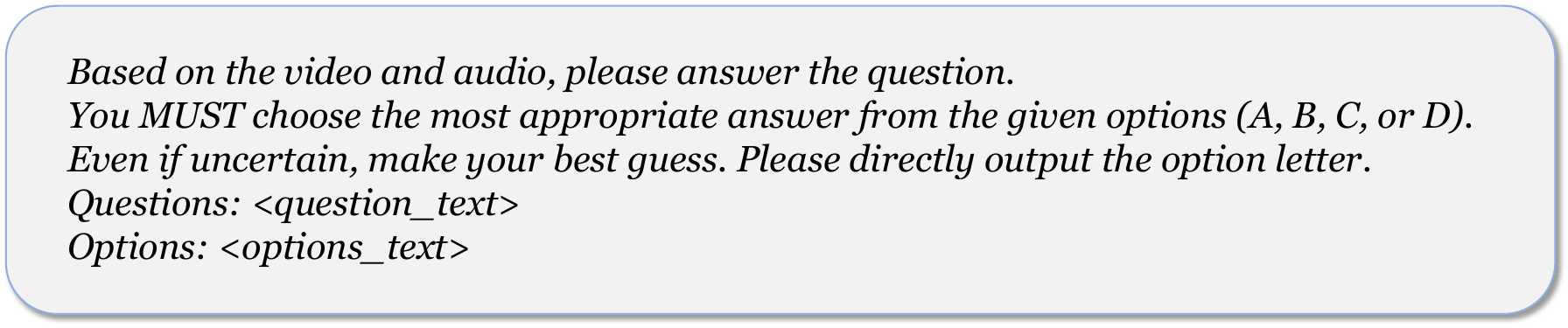} 
    \vspace{-0.5cm}
  
    \label{fig:prompt_av}
\end{figure}

\clearpage
\section{Failure case study}
\label{appendix_h}

\begin{figure}[H] 
    \centering 
    \includegraphics[width=1.0\textwidth]{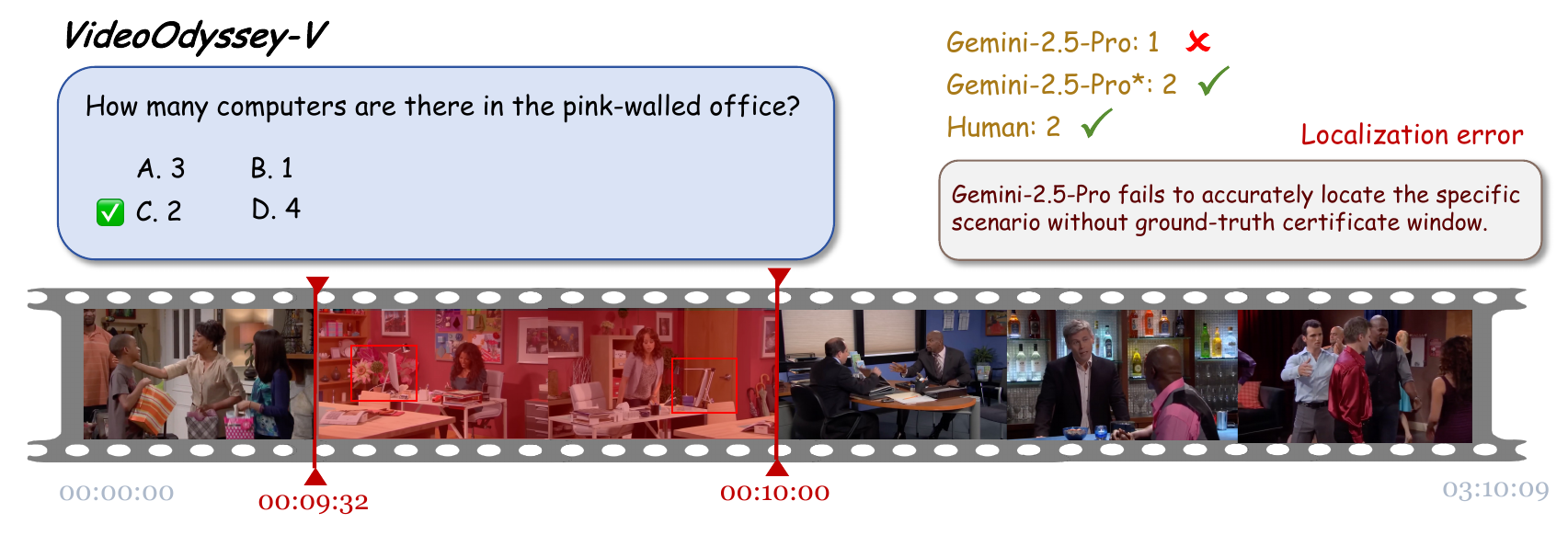} 
    \caption{Case of localization error. * indicates using the certificate window, whereas no asterisk denotes the default entire-video input.}
    \label{fig:failure_case_localization}
\end{figure}

\begin{figure}[H] 
    \centering 
    \includegraphics[width=1.0\textwidth]{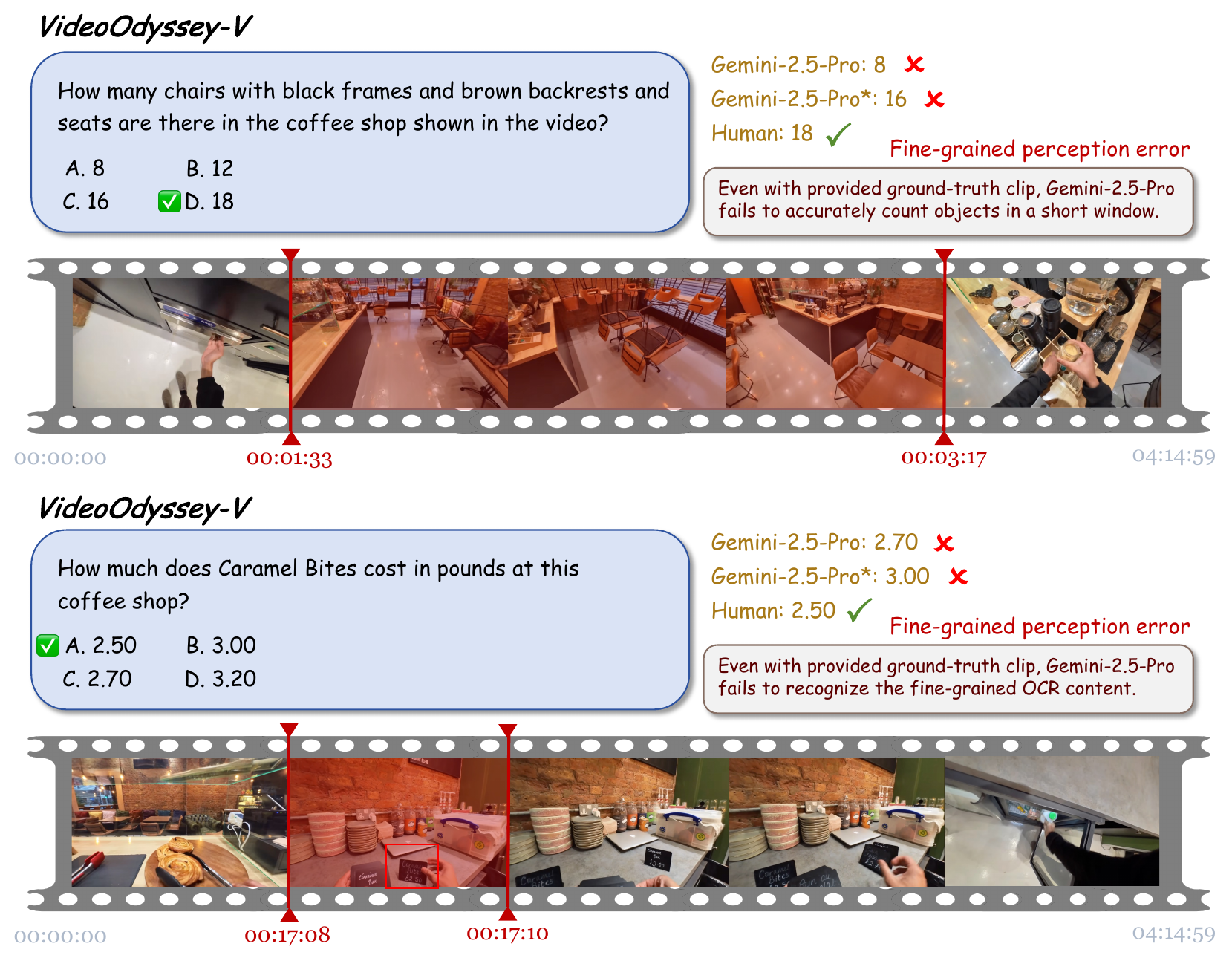} 
    \vspace{-0.5cm}
    \caption{Cases of fain-grained perception error. * indicates using the certificate window, whereas no asterisk denotes the default entire-video input.}
    \label{fig:failure_case_fain_grained_perception}
    \vspace{-0.4cm}
\end{figure}

\begin{figure}[H] 
    \centering 
    \includegraphics[width=1.0\textwidth]{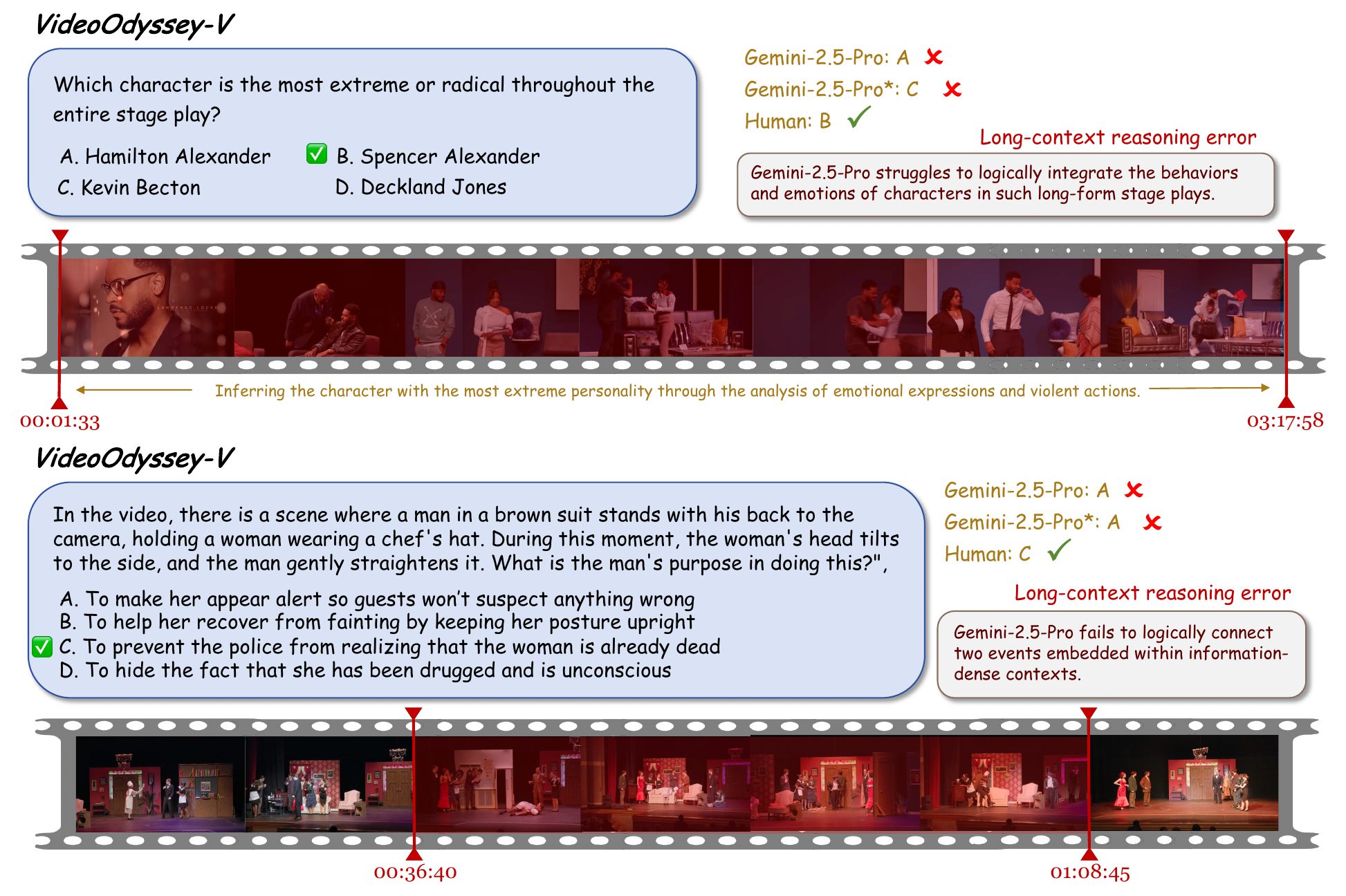} 
    \caption{Cases of long-context reasoning error. * indicates using the certificate window, whereas no asterisk denotes the default entire-video input.}
    \label{fig:failure_case_long_context}
\end{figure}

\begin{figure}[H] 
    \centering 
    \includegraphics[width=1.0\textwidth]{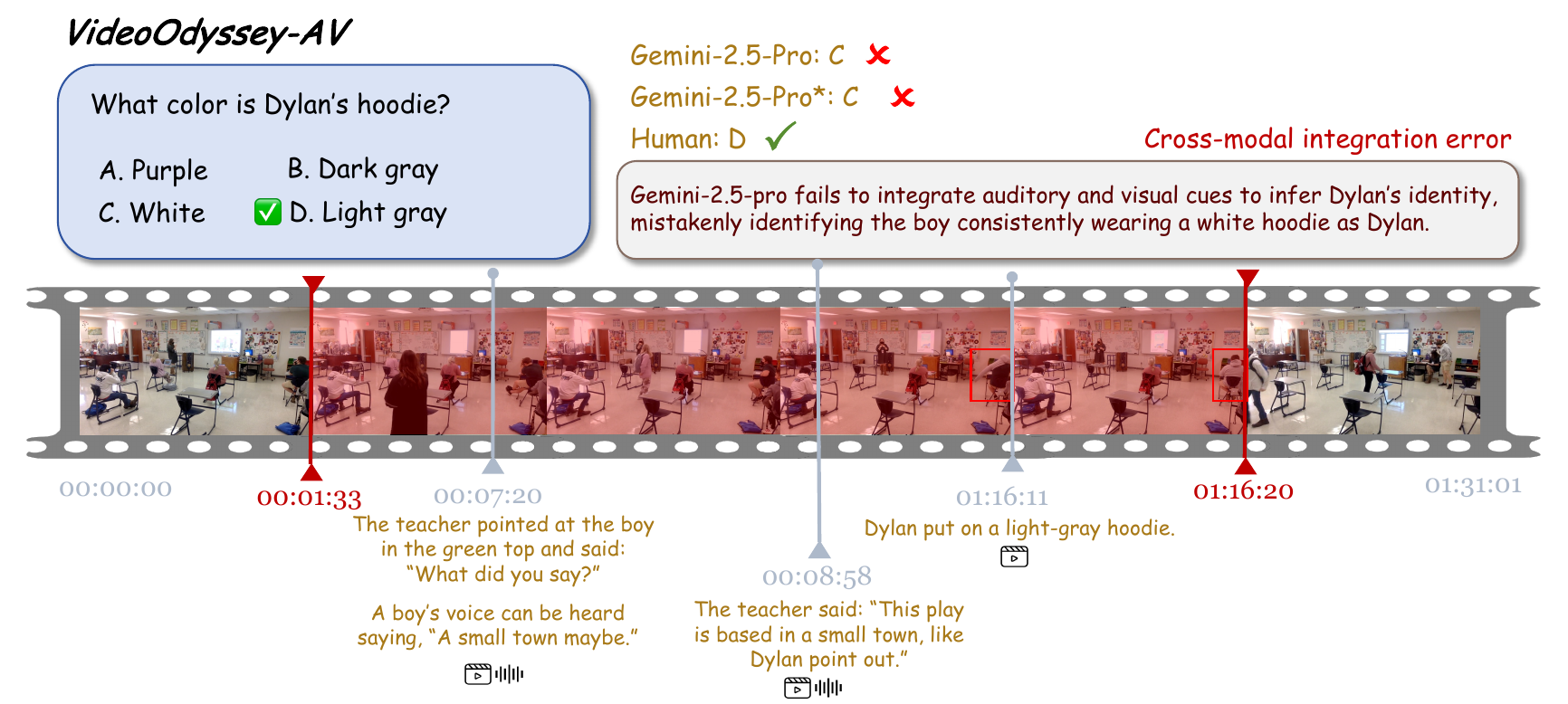} 
    \caption{Case of cross-modal integration error. * indicates using the certificate window, whereas no asterisk denotes the default entire-video input.}
    \label{fig:failure_case_cross_modal}
\end{figure}

\begin{figure}[H] 
\vspace{-0.2cm}
    \centering 
    \includegraphics[width=1.0\textwidth]{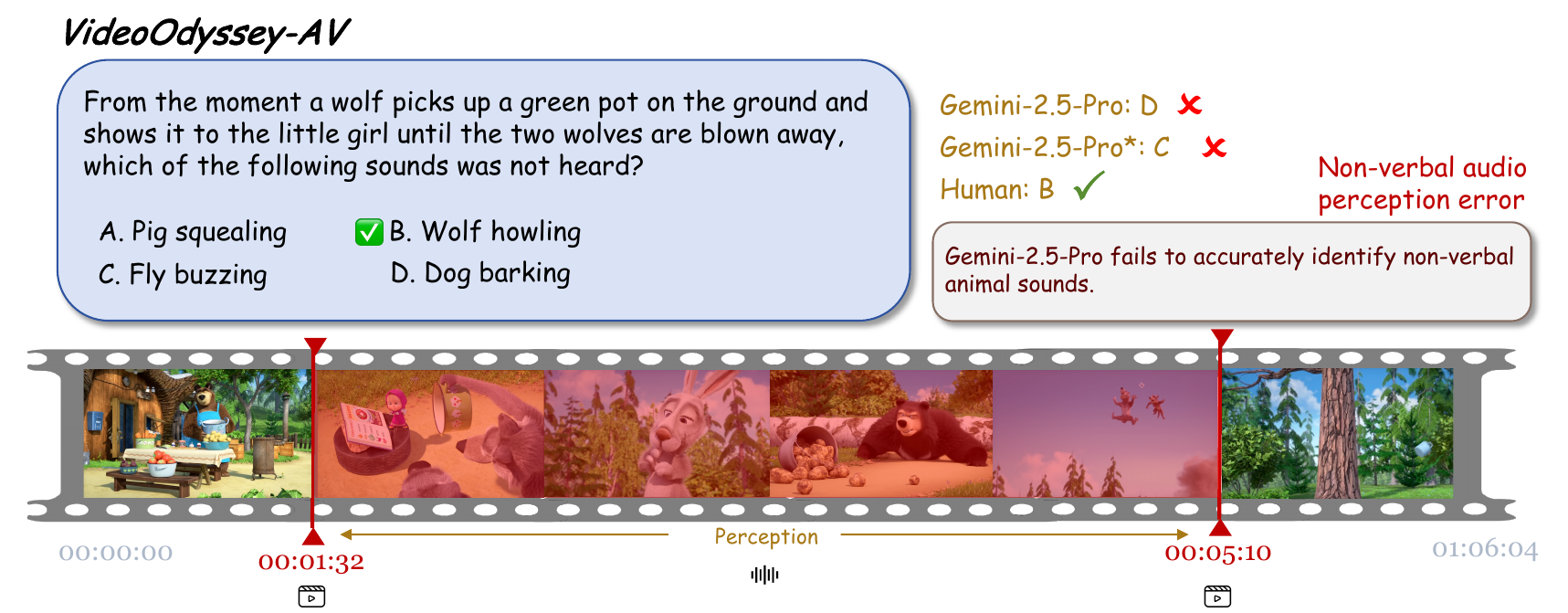} 
    \vspace{-0.5cm}
    \caption{Case of non-verbal audio perception error. * indicates using the certificate window, whereas no asterisk denotes the default entire-video input.}
    \label{fig:failure_case_non_verbal}
\end{figure}

We conduct a qualitative analysis of failure cases to shed light on model limitations. We uses Gemini-2.5-Pro as an example and presents five distinct error types: localization error (Fig. \ref{fig:failure_case_localization}), fine-grained perception error (Fig. \ref{fig:failure_case_fain_grained_perception}), long-context reasoning error (Fig. \ref{fig:failure_case_long_context}), cross-modal integration error (Fig. \ref{fig:failure_case_cross_modal}), and non-verbal audio perception error (Fig. \ref{fig:failure_case_non_verbal}).

\section{Limitations and broader impacts}
\label{appendix_i}

Despite our efforts, \textit{VideoOdyssey} presents two primary limitations. First, due to the severe cognitive load and expert-level temporal reasoning required to annotate ultra-long contexts, scaling the dataset is exceptionally labor-intensive, resulting in a smaller size compared to short-video benchmarks. Second, our human evaluation captures overall accuracy without recording the exact time spent per question. Capturing these fine-grained temporal metrics could reveal how human seeking behavior correlates with continuous certificate length, offering valuable inspiration for designing efficient video reasoning agents.

Advancing ultra-long video understanding offers significant societal benefits, such as improving educational analysis and developing assistive technologies for the visually impaired. However, enhanced capabilities in continuous spatial-temporal reasoning also raise ethical concerns regarding potential misuse in automated mass surveillance and privacy violations. We encourage the research community to establish robust frameworks to ensure these long-context models are deployed responsibly.

\newpage
\section*{NeurIPS Paper Checklist}

\begin{enumerate}

\item {\bf Claims}
    \item[] Question: Do the main claims made in the abstract and introduction accurately reflect the paper's contributions and scope?
    \item[] Answer: \answerYes{} 
    \item[] Justification: The abstract and introduction accurately state our core contributions, including the VideoOdyssey benchmark, the continuous certificate length metric and insights from benchmarking. These claims are directly supported by Sec \ref{videoodyssey_construction} and \ref{experiments}.
    \item[] Guidelines:
    \begin{itemize}
        \item The answer \answerNA{} means that the abstract and introduction do not include the claims made in the paper.
        \item The abstract and/or introduction should clearly state the claims made, including the contributions made in the paper and important assumptions and limitations. A \answerNo{} or \answerNA{} answer to this question will not be perceived well by the reviewers. 
        \item The claims made should match theoretical and experimental results, and reflect how much the results can be expected to generalize to other settings. 
        \item It is fine to include aspirational goals as motivation as long as it is clear that these goals are not attained by the paper. 
    \end{itemize}

\item {\bf Limitations}
    \item[] Question: Does the paper discuss the limitations of the work performed by the authors?
    \item[] Answer: \answerYes{} 
    \item[] Justification: Due to page limits in the main text, a detailed discussion of the limitations is provided in appendix \ref{appendix_i}.
    \item[] Guidelines:
    \begin{itemize}
        \item The answer \answerNA{} means that the paper has no limitation while the answer \answerNo{} means that the paper has limitations, but those are not discussed in the paper. 
        \item The authors are encouraged to create a separate ``Limitations'' section in their paper.
        \item The paper should point out any strong assumptions and how robust the results are to violations of these assumptions (e.g., independence assumptions, noiseless settings, model well-specification, asymptotic approximations only holding locally). The authors should reflect on how these assumptions might be violated in practice and what the implications would be.
        \item The authors should reflect on the scope of the claims made, e.g., if the approach was only tested on a few datasets or with a few runs. In general, empirical results often depend on implicit assumptions, which should be articulated.
        \item The authors should reflect on the factors that influence the performance of the approach. For example, a facial recognition algorithm may perform poorly when image resolution is low or images are taken in low lighting. Or a speech-to-text system might not be used reliably to provide closed captions for online lectures because it fails to handle technical jargon.
        \item The authors should discuss the computational efficiency of the proposed algorithms and how they scale with dataset size.
        \item If applicable, the authors should discuss possible limitations of their approach to address problems of privacy and fairness.
        \item While the authors might fear that complete honesty about limitations might be used by reviewers as grounds for rejection, a worse outcome might be that reviewers discover limitations that aren't acknowledged in the paper. The authors should use their best judgment and recognize that individual actions in favor of transparency play an important role in developing norms that preserve the integrity of the community. Reviewers will be specifically instructed to not penalize honesty concerning limitations.
    \end{itemize}

\item {\bf Theory assumptions and proofs}
    \item[] Question: For each theoretical result, does the paper provide the full set of assumptions and a complete (and correct) proof?
    \item[] Answer: \answerNA{} 
    \item[] Justification: This paper introduces a benchmark for evaluating MLLMS and does not propose new theoretical theorems or mathematical proofs.
    \item[] Guidelines:
    \begin{itemize}
        \item The answer \answerNA{} means that the paper does not include theoretical results. 
        \item All the theorems, formulas, and proofs in the paper should be numbered and cross-referenced.
        \item All assumptions should be clearly stated or referenced in the statement of any theorems.
        \item The proofs can either appear in the main paper or the supplemental material, but if they appear in the supplemental material, the authors are encouraged to provide a short proof sketch to provide intuition. 
        \item Inversely, any informal proof provided in the core of the paper should be complemented by formal proofs provided in appendix or supplemental material.
        \item Theorems and Lemmas that the proof relies upon should be properly referenced. 
    \end{itemize}

    \item {\bf Experimental result reproducibility}
    \item[] Question: Does the paper fully disclose all the information needed to reproduce the main experimental results of the paper to the extent that it affects the main claims and/or conclusions of the paper (regardless of whether the code and data are provided or not)?
    \item[] Answer: \answerYes{} 
    \item[] Justification: Detailed evaluation protocols are described in the main text, and we provide the dataset, prompt templates, and evaluation scripts in the anonymous GitHub repository to ensure full reproducibility.
    \item[] Guidelines:
    \begin{itemize}
        \item The answer \answerNA{} means that the paper does not include experiments.
        \item If the paper includes experiments, a \answerNo{} answer to this question will not be perceived well by the reviewers: Making the paper reproducible is important, regardless of whether the code and data are provided or not.
        \item If the contribution is a dataset and\slash or model, the authors should describe the steps taken to make their results reproducible or verifiable. 
        \item Depending on the contribution, reproducibility can be accomplished in various ways. For example, if the contribution is a novel architecture, describing the architecture fully might suffice, or if the contribution is a specific model and empirical evaluation, it may be necessary to either make it possible for others to replicate the model with the same dataset, or provide access to the model. In general. releasing code and data is often one good way to accomplish this, but reproducibility can also be provided via detailed instructions for how to replicate the results, access to a hosted model (e.g., in the case of a large language model), releasing of a model checkpoint, or other means that are appropriate to the research performed.
        \item While NeurIPS does not require releasing code, the conference does require all submissions to provide some reasonable avenue for reproducibility, which may depend on the nature of the contribution. For example
        \begin{enumerate}
            \item If the contribution is primarily a new algorithm, the paper should make it clear how to reproduce that algorithm.
            \item If the contribution is primarily a new model architecture, the paper should describe the architecture clearly and fully.
            \item If the contribution is a new model (e.g., a large language model), then there should either be a way to access this model for reproducing the results or a way to reproduce the model (e.g., with an open-source dataset or instructions for how to construct the dataset).
            \item We recognize that reproducibility may be tricky in some cases, in which case authors are welcome to describe the particular way they provide for reproducibility. In the case of closed-source models, it may be that access to the model is limited in some way (e.g., to registered users), but it should be possible for other researchers to have some path to reproducing or verifying the results.
        \end{enumerate}
    \end{itemize}

\item {\bf Open access to data and code}
    \item[] Question: Does the paper provide open access to the data and code, with sufficient instructions to faithfully reproduce the main experimental results, as described in supplemental material?
    \item[] Answer: \answerYes{} 
    \item[] Justification: All dataset annotations, evaluation scripts, and detailed instructions for reproducing our results are provided via our anonymized project website and GitHub repository.
    \item[] Guidelines:
    \begin{itemize}
        \item The answer \answerNA{} means that paper does not include experiments requiring code.
        \item Please see the NeurIPS code and data submission guidelines (\url{https://neurips.cc/public/guides/CodeSubmissionPolicy}) for more details.
        \item While we encourage the release of code and data, we understand that this might not be possible, so \answerNo{} is an acceptable answer. Papers cannot be rejected simply for not including code, unless this is central to the contribution (e.g., for a new open-source benchmark).
        \item The instructions should contain the exact command and environment needed to run to reproduce the results. See the NeurIPS code and data submission guidelines (\url{https://neurips.cc/public/guides/CodeSubmissionPolicy}) for more details.
        \item The authors should provide instructions on data access and preparation, including how to access the raw data, preprocessed data, intermediate data, and generated data, etc.
        \item The authors should provide scripts to reproduce all experimental results for the new proposed method and baselines. If only a subset of experiments are reproducible, they should state which ones are omitted from the script and why.
        \item At submission time, to preserve anonymity, the authors should release anonymized versions (if applicable).
        \item Providing as much information as possible in supplemental material (appended to the paper) is recommended, but including URLs to data and code is permitted.
    \end{itemize}

\item {\bf Experimental setting/details}
    \item[] Question: Does the paper specify all the training and test details (e.g., data splits, hyperparameters, how they were chosen, type of optimizer) necessary to understand the results?
    \item[] Answer: \answerYes{} 
    \item[] Justification: Since our work focuses on evaluation rather than training, all necessary inference details—including model versions, prompt templates, and evaluation metrics—are fully specified in the main text and Appendix.
    \item[] Guidelines:
    \begin{itemize}
        \item The answer \answerNA{} means that the paper does not include experiments.
        \item The experimental setting should be presented in the core of the paper to a level of detail that is necessary to appreciate the results and make sense of them.
        \item The full details can be provided either with the code, in appendix, or as supplemental material.
    \end{itemize}

\item {\bf Experiment statistical significance}
    \item[] Question: Does the paper report error bars suitably and correctly defined or other appropriate information about the statistical significance of the experiments?
    \item[] Answer: \answerNo{} 
    \item[] Justification: As is standard practice for MLLM benchmark evaluations, we report absolute accuracy scores on a fixed test set using deterministic or low-temperature generation settings, where variance across runs is negligible and error bars are not traditionally required.
    \item[] Guidelines:
    \begin{itemize}
        \item The answer \answerNA{} means that the paper does not include experiments.
        \item The authors should answer \answerYes{} if the results are accompanied by error bars, confidence intervals, or statistical significance tests, at least for the experiments that support the main claims of the paper.
        \item The factors of variability that the error bars are capturing should be clearly stated (for example, train/test split, initialization, random drawing of some parameter, or overall run with given experimental conditions).
        \item The method for calculating the error bars should be explained (closed form formula, call to a library function, bootstrap, etc.)
        \item The assumptions made should be given (e.g., Normally distributed errors).
        \item It should be clear whether the error bar is the standard deviation or the standard error of the mean.
        \item It is OK to report 1-sigma error bars, but one should state it. The authors should preferably report a 2-sigma error bar than state that they have a 96\% CI, if the hypothesis of Normality of errors is not verified.
        \item For asymmetric distributions, the authors should be careful not to show in tables or figures symmetric error bars that would yield results that are out of range (e.g., negative error rates).
        \item If error bars are reported in tables or plots, the authors should explain in the text how they were calculated and reference the corresponding figures or tables in the text.
    \end{itemize}

\item {\bf Experiments compute resources}
    \item[] Question: For each experiment, does the paper provide sufficient information on the computer resources (type of compute workers, memory, time of execution) needed to reproduce the experiments?
    \item[] Answer: \answerYes{} 
    \item[] Justification: We detail the computer resources in Sec \ref{experiments}.
    \item[] Guidelines:
    \begin{itemize}
        \item The answer \answerNA{} means that the paper does not include experiments.
        \item The paper should indicate the type of compute workers CPU or GPU, internal cluster, or cloud provider, including relevant memory and storage.
        \item The paper should provide the amount of compute required for each of the individual experimental runs as well as estimate the total compute. 
        \item The paper should disclose whether the full research project required more compute than the experiments reported in the paper (e.g., preliminary or failed experiments that didn't make it into the paper). 
    \end{itemize}
    
\item {\bf Code of ethics}
    \item[] Question: Does the research conducted in the paper conform, in every respect, with the NeurIPS Code of Ethics \url{https://neurips.cc/public/EthicsGuidelines}?
    \item[] Answer: \answerYes{} 
    \item[] Justification: The VideoOdyssey dataset is sourced from publicly available YouTube videos and does not contain content that violates personal privacy. Furthermore, we have established a takedown policy on our project website to promptly address any potential concerns.
    \item[] Guidelines:
    \begin{itemize}
        \item The answer \answerNA{} means that the authors have not reviewed the NeurIPS Code of Ethics.
        \item If the authors answer \answerNo, they should explain the special circumstances that require a deviation from the Code of Ethics.
        \item The authors should make sure to preserve anonymity (e.g., if there is a special consideration due to laws or regulations in their jurisdiction).
    \end{itemize}

\item {\bf Broader impacts}
    \item[] Question: Does the paper discuss both potential positive societal impacts and negative societal impacts of the work performed?
    \item[] Answer: \answerYes{} 
    \item[] Justification: 
    A dedicated discussion on the broader societal impacts is provided in the appendix \ref{appendix_i}.
    \item[] Guidelines:
    \begin{itemize}
        \item The answer \answerNA{} means that there is no societal impact of the work performed.
        \item If the authors answer \answerNA{} or \answerNo, they should explain why their work has no societal impact or why the paper does not address societal impact.
        \item Examples of negative societal impacts include potential malicious or unintended uses (e.g., disinformation, generating fake profiles, surveillance), fairness considerations (e.g., deployment of technologies that could make decisions that unfairly impact specific groups), privacy considerations, and security considerations.
        \item The conference expects that many papers will be foundational research and not tied to particular applications, let alone deployments. However, if there is a direct path to any negative applications, the authors should point it out. For example, it is legitimate to point out that an improvement in the quality of generative models could be used to generate Deepfakes for disinformation. On the other hand, it is not needed to point out that a generic algorithm for optimizing neural networks could enable people to train models that generate Deepfakes faster.
        \item The authors should consider possible harms that could arise when the technology is being used as intended and functioning correctly, harms that could arise when the technology is being used as intended but gives incorrect results, and harms following from (intentional or unintentional) misuse of the technology.
        \item If there are negative societal impacts, the authors could also discuss possible mitigation strategies (e.g., gated release of models, providing defenses in addition to attacks, mechanisms for monitoring misuse, mechanisms to monitor how a system learns from feedback over time, improving the efficiency and accessibility of ML).
    \end{itemize}
    
\item {\bf Safeguards}
    \item[] Question: Does the paper describe safeguards that have been put in place for responsible release of data or models that have a high risk for misuse (e.g., pre-trained language models, image generators, or scraped datasets)?
    \item[] Answer: \answerYes{} 
    \item[] Justification: To mitigate potential risks associated with sourced datasets, we implemented safeguards by manually filtering videos for sensitive content and establishing a formal takedown policy on our project website.
    \item[] Guidelines:
    \begin{itemize}
        \item The answer \answerNA{} means that the paper poses no such risks.
        \item Released models that have a high risk for misuse or dual-use should be released with necessary safeguards to allow for controlled use of the model, for example by requiring that users adhere to usage guidelines or restrictions to access the model or implementing safety filters. 
        \item Datasets that have been scraped from the Internet could pose safety risks. The authors should describe how they avoided releasing unsafe images.
        \item We recognize that providing effective safeguards is challenging, and many papers do not require this, but we encourage authors to take this into account and make a best faith effort.
    \end{itemize}

\item {\bf Licenses for existing assets}
    \item[] Question: Are the creators or original owners of assets (e.g., code, data, models), used in the paper, properly credited and are the license and terms of use explicitly mentioned and properly respected?
    \item[] Answer: \answerYes{} 
    \item[] Justification: All evaluated models and existing tools are properly cited. Furthermore, our dataset is released under the CC BY-NC 4.0 license, utilizing YouTube videos strictly under academic fair use principles.
    \item[] Guidelines:
    \begin{itemize}
        \item The answer \answerNA{} means that the paper does not use existing assets.
        \item The authors should cite the original paper that produced the code package or dataset.
        \item The authors should state which version of the asset is used and, if possible, include a URL.
        \item The name of the license (e.g., CC-BY 4.0) should be included for each asset.
        \item For scraped data from a particular source (e.g., website), the copyright and terms of service of that source should be provided.
        \item If assets are released, the license, copyright information, and terms of use in the package should be provided. For popular datasets, \url{paperswithcode.com/datasets} has curated licenses for some datasets. Their licensing guide can help determine the license of a dataset.
        \item For existing datasets that are re-packaged, both the original license and the license of the derived asset (if it has changed) should be provided.
        \item If this information is not available online, the authors are encouraged to reach out to the asset's creators.
    \end{itemize}

\item {\bf New assets}
    \item[] Question: Are new assets introduced in the paper well documented and is the documentation provided alongside the assets?
    \item[] Answer: \answerYes{} 
    \item[] Justification: Comprehensive documentation for the VideoOdyssey benchmark, including dataset statistics, evaluation protocols, and licensing, is provided on our anonymized project website and GitHub repository.
    \item[] Guidelines:
    \begin{itemize}
        \item The answer \answerNA{} means that the paper does not release new assets.
        \item Researchers should communicate the details of the dataset\slash code\slash model as part of their submissions via structured templates. This includes details about training, license, limitations, etc. 
        \item The paper should discuss whether and how consent was obtained from people whose asset is used.
        \item At submission time, remember to anonymize your assets (if applicable). You can either create an anonymized URL or include an anonymized zip file.
    \end{itemize}

\item {\bf Crowdsourcing and research with human subjects}
    \item[] Question: For crowdsourcing experiments and research with human subjects, does the paper include the full text of instructions given to participants and screenshots, if applicable, as well as details about compensation (if any)? 
    \item[] Answer: \answerYes{} 
    \item[] Justification: Guidelines for dataset annotation are detailed in Sec \ref{videoodyssey_construction}, while the specific instructions for human performance evaluators are provided in the appendix \ref{appendix_d}. Both annotators and participants were fairly compensated in accordance with local labor standards.
    \item[] Guidelines:
    \begin{itemize}
        \item The answer \answerNA{} means that the paper does not involve crowdsourcing nor research with human subjects.
        \item Including this information in the supplemental material is fine, but if the main contribution of the paper involves human subjects, then as much detail as possible should be included in the main paper. 
        \item According to the NeurIPS Code of Ethics, workers involved in data collection, curation, or other labor should be paid at least the minimum wage in the country of the data collector. 
    \end{itemize}

\item {\bf Institutional review board (IRB) approvals or equivalent for research with human subjects}
    \item[] Question: Does the paper describe potential risks incurred by study participants, whether such risks were disclosed to the subjects, and whether Institutional Review Board (IRB) approvals (or an equivalent approval/review based on the requirements of your country or institution) were obtained?
    \item[] Answer: \answerNo{} 
    \item[] Justification: The human evaluation strictly involves standard video annotation and question-answering tasks without physical or psychological risks, which is typically exempt from formal IRB approval.
    \item[] Guidelines:
    \begin{itemize}
        \item The answer \answerNA{} means that the paper does not involve crowdsourcing nor research with human subjects.
        \item Depending on the country in which research is conducted, IRB approval (or equivalent) may be required for any human subjects research. If you obtained IRB approval, you should clearly state this in the paper. 
        \item We recognize that the procedures for this may vary significantly between institutions and locations, and we expect authors to adhere to the NeurIPS Code of Ethics and the guidelines for their institution. 
        \item For initial submissions, do not include any information that would break anonymity (if applicable), such as the institution conducting the review.
    \end{itemize}

\item {\bf Declaration of LLM usage}
    \item[] Question: Does the paper describe the usage of LLMs if it is an important, original, or non-standard component of the core methods in this research? Note that if the LLM is used only for writing, editing, or formatting purposes and does \emph{not} impact the core methodology, scientific rigor, or originality of the research, declaration is not required.
    \item[] Answer: \answerYes{} 
    \item[] Justification: We utilized Large Language Models to assist in controlling the quality of dataset annotations, the specific methodology and implementation of which are detailed in Sec \ref{videoodyssey_quality_control}.
    \item[] Guidelines:
    \begin{itemize}
        \item The answer \answerNA{} means that the core method development in this research does not involve LLMs as any important, original, or non-standard components.
        \item Please refer to our LLM policy in the NeurIPS handbook for what should or should not be described.
    \end{itemize}

\end{enumerate}

\end{document}